\documentclass{article}

\usepackage{xcolor}
\usepackage{colortbl}
\usepackage{microtype}
\usepackage{graphicx}
\usepackage{subfigure}
\usepackage{booktabs} 
\usepackage{multirow}

\usepackage{hyperref}

\usepackage{adjustbox}

\usepackage[accepted]{icml2024}

\usepackage{amsmath}
\usepackage{amssymb}
\usepackage{mathtools}
\usepackage{amsthm}

\usepackage[capitalize,noabbrev]{cleveref}

\theoremstyle{plain}

\theoremstyle{definition}

\theoremstyle{remark}

\usepackage[textsize=tiny]{todonotes}

\icmltitlerunning{\textsc{Router}\textsc{Bench}: A Benchmark for Multi-LLM Routing System}

\begin{document}

\twocolumn[
\icmltitle{\textsc{Router}\textsc{Bench}: A Benchmark for Multi-LLM Routing System}



\icmlsetsymbol{equal}{*}
\icmlsetsymbol{lead}{$\dagger$}


\begin{icmlauthorlist}
\icmlauthor{Qitian Jason Hu}{mars}
\icmlauthor{Jacob Bieker}{mars}
\icmlauthor{Xiuyu Li}{ucb}
\icmlauthor{Nan Jiang}{ucsd}
\icmlauthor{Benjamin Keigwin}{mars}
\icmlauthor{Gaurav Ranganath}{mars}
\icmlauthor{Kurt Keutzer}{ucb}
\icmlauthor{Shriyash Kaustubh Upadhyay}{mars}
\end{icmlauthorlist}

\icmlaffiliation{mars}{Martian}
\icmlaffiliation{ucb}{UC Berkeley}
\icmlaffiliation{ucsd}{UC San Diego}

\icmlcorrespondingauthor{Qitian Jason Hu}{jason@withmartian.com}

\icmlkeywords{Machine Learning, ICML}

\vskip 0.3in
]



\printAffiliationsAndNotice{}  

\newcommand{\sect}[1]{Section~\ref{#1}}
\newcommand{\ssect}[1]{\S~\ref{#1}}
\newcommand{\append}[1]{Appendix~\ref{#1}}
\newcommand{\eqn}[1]{Equation~\ref{#1}}
\newcommand{\fig}[1]{Figure~\ref{#1}}
\newcommand{\tbl}[1]{Table~\ref{#1}}
\newcommand{\algo}[1]{Algorithm~\ref{#1}}
\newcommand{\myparagraph}[1]{\noindent \textbf{#1}}
\newcommand{\na}{---}
\newcommand{\ourcell}{\cellcolor[rgb]{1,0.808,0.576}}
\newcommand\hc{ \rowcolor{teal!15}}

\renewcommand{\todo}[1]{\textcolor{red}{[TODO: #1]}}
\newcommand{\jason}[1]{\textcolor{blue}{[Jason: #1]}}
\newcommand{\yash}[1]{\textcolor{cyan}{[Yash: #1]}}
\newcommand{\xiuyu}[1]{\textcolor{magenta}{[Xiuyu: #1]}}
\newcommand{\will}[1]{\textcolor{orange}{[Will: #1]}}
\newcommand{\update}[1]{\textcolor{blue}{#1}}

\newcommand{\dataset}{\mbox{\textsc{RouterBench}}}  %

\begin{abstract}


As the range of applications for Large Language Models (LLMs) continues to grow, the demand for effective serving solutions becomes increasingly critical. Despite the versatility of LLMs, no single model can optimally address all tasks and applications, particularly when balancing performance with cost. This limitation has led to the development of LLM routing systems, which combine the strengths of various models to overcome the constraints of individual LLMs. Yet, the absence of a standardized benchmark for evaluating the performance of LLM routers hinders progress in this area. To bridge this gap, we present~\dataset, a novel evaluation framework designed to systematically assess the efficacy of LLM routing systems, along with a comprehensive dataset comprising
over 405k inference outcomes from representative LLMs to support the development of routing strategies. We further propose a theoretical framework for LLM routing, and deliver a comparative analysis of various routing approaches through~\dataset, highlighting their potentials and limitations within our evaluation framework. This work not only formalizes and advances the development of LLM routing systems but also sets a standard for their assessment, paving the way for more accessible and economically viable LLM deployments. The code and data are available at https://github.com/withmartian/routerbench.

\end{abstract}

\begin{figure*}[ht]
    \centering
    \includegraphics[width=\textwidth]{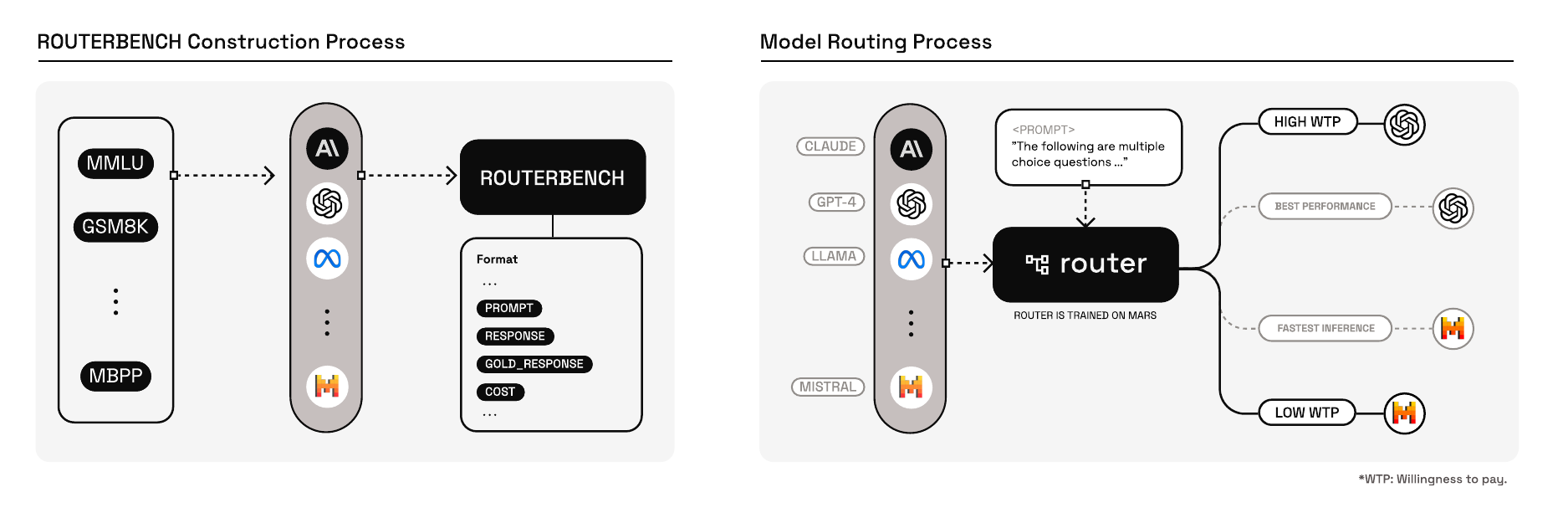}
    \vspace{-20pt}
    \caption{ \textbf{Left}: The \dataset~Construction Process integrates eight datasets with eleven distinct models to develop \dataset. Detailed format can be found in Appendix \ref{appendix:data_entry}. \textbf{Right}: The Model Routing Process shows the method of routing prompts through a router to various LLMs based on specific requests, demonstrating the dynamic allocation of resources.}
    \vspace{-10pt}
    \label{fig:teaser}
\end{figure*}

\section{Introduction} 






Large Language Models (LLMs) have exhibited remarkable capabilities in addressing a wide range of tasks across academic and industrial scenarios \cite{bubeck2023sparks}. This has motivated both researchers and practitioners to introduce new LLMs, designed for both generic and specialized use cases, on a near-daily basis \footnote{As of January 16th, 2024, there are 469,848 models listed on huggingface.com}. However, the proliferation of LLMs presents a challenge for LLM application builders to identify the most suitable model for their applications. While some proprietary models, such as GPT-4, are distinguished by their superior performance, they often incur high economic costs due to the high API prices. 


Many prior works focus on improving the capabilities of individual LLMs while maintaining low costs. Techniques such as prompting~\cite{cot}, quantization~\cite{lin2023awq, kim2023squeezellm}, and system optimization~\cite{kwon2023efficient} may reduce a single model's serving cost, yet with new models emerging daily, these approaches may not remain feasible or scalable in long term. Moreover, the diversity of choices of LLMs available at various price and performance tiers can be daunting for users attempting to select and optimize an appropriate model\footnote{As of January 29th, 2024, there are 22,177 language models with 7 billion parameters listed on huggingface.com}.


An alternative solution aims to select to optimal LLM for each input through "routing." \cite{frugalgpt, llmbenchmark, flyswat}. Routing offers several advantages over single-LLM optimization. First, it is a lightweight process, which treats each LLM as an input-output black box, avoiding the need to delve into intricate infrastructure details, thus making it flexible and broadly applicable. 
Second, routing systems benefit from the diversity of LLMs,  while single-LLM methods may struggle to keep pace with the expanding LLM landscape.
Lastly, while single-LLM strategies often face a compromise between performance and other factors such as per-token costs, routing systems adeptly balance a spectrum of user demands.

The rise in routing-related research has improved cost efficiency, enhanced performance, and broadened accessibility~\cite{frugalgpt, orchestrallm, routingtoexpert}. Despite these advances, a comprehensive benchmark for evaluating routing techniques remains absent. 
We introduce \dataset, the first comprehensive benchmark designed specifically for assessing router mechanisms in terms of inference dollar cost and performance. \dataset~encompasses a diverse array of tasks and domains, with pre-generated LLM response and quality metrics, on which different routing mechanisms can be efficiently tested without inference. Our experiments revealed that while some previous routing mechanisms have difficulty generalizing to complex tasks and up-to-date models, there are several promising fields on which even simple routing demonstrated outstanding performance. 


In conclusion, we present the following key contributions:
\begin{enumerate}
    \item We have developed a comprehensive benchmark for LLM routing covering major tasks for LLMs, which includes a wide range of both open-source and proprietary models. \dataset~enables efficient training and testing of model routers without inference, and can be flexibly extended to cover new tasks and models.
    \item We introduce a theoretical framework designed to assess the efficacy of routers across several metrics, with a particular emphasis on inference cost (expressed in dollars) and performance. This framework includes mathematical formulations that enable the seamless integration and comparative analysis of various routers and LLMs.
    \item We evaluate the efficiency of routing strategies across a broad range of tasks. Our results provide insights into the performance of various routers in different contexts and demonstrate that the monetary costs of LLM services can routinely vary by factors of $2$-$5\times$ for comparable levels of performance. This underscores the significance and utility of our benchmark, highlighting promising areas for future enhancements.
\end{enumerate}


\section{Related Work} 
\label{2_related_work}


Various strategies have been proposed to optimize the cost and performance of current LLMs. We provide an overview of them with a focus on routing-related approaches. 

\myparagraph{Single LLM Enhancement}
Fine-tuning is used to improve models for specific tasks, which requires additional training and domain-specific data~\cite{rafailov2023direct}. Prompting mechanisms like Chain-of-Thought (CoT)~\cite{cot, leasttomost, wang2023selfconsistency} and Tree of Thoughts (ToT)~\cite{tot} could bolster LLM performance without additional fine-tuning. Mixture-of-Experts (MoE)~\cite{dMoE, outrageously, switch, glam, shenmoeit, si2023getting} is another line of work that explores routing within the model to enhance performance efficiently, which contains specialized "experts" and routes the input to the best expert. Nevertheless, these single-LLM enhancements are usually model and scenario specific, and could not benefit from the explosion in the number of LLMs. 


\myparagraph{LLM Synthesis}
Beyond single LLM approaches, LLM synthesis utilizes the ensemble of multiple LLMs, integrating their outputs into an enhanced final result~\cite{llmblender}. Another approach has shown that a strategic combination of smaller models can match or even outperform larger models~\cite{blending}. However, these methods require at least two steps: text generation and synthesis, which increases costs and latency, creating challenges to applying this approach in production.

\myparagraph{Routing}
Unlike LLM Synthesis, routing can select the suitable model for specific input without performing inference on every candidate model. Routing can be classified into two categories, non-predictive routing and predictive routing. Non-predictive routing strategies retrieve outputs from LLMs and directly pick one without a model-assisted synthesis step. FrugalGPT~\cite{frugalgpt} presents an inaugural application of this type of strategy, which employs a generation judger that assesses the quality of responses from various LLMs to a given query, invoking LLMs sequentially until an answer meets a predefined quality threshold. Several studies~\cite{automix, llmcascades, orchestrallm} also have explored systems that integrate small language models with LLMs. Another methodology involves a layered inference framework, re-routing more complex queries to an advanced model for improved results~\cite{tabi}. 
Predictive routing selects the optimal LLM without requiring to evaluate the output. One line of research has implemented routers utilizing supervised learning algorithms~\cite{llmbenchmark}, while some others use reward model-based techniques~\cite{tryage, routingtoexpert}. Furthermore, meta-model, trained on inputs along with a model-specific token to predict the performance score, represents another approach to determining the most appropriate LLM for use~\cite{flyswat}. In short, predictive routers could bring substantial cost and performance improvement without sacrificing latency, with a number of early works dedicated to this field.

While many routers currently exist, a systematic benchmark for their evaluation has been lacking. Our work aims to address this issue and introduce a benchmark for router evaluation.

\section{Math Formulation for Router Evaluation} 

The primary challenge in assessing the performance of routing systems lies in balancing two conflicting objectives: maximizing efficiency and minimizing cost. To effectively compare routers, we have developed a framework that captures the multi-faceted nature with one metric. 

\subsection{Setup and Basic Operations}

Consider a set of models $L = \{LLM_1,\ldots, LLM_m\}$ a dataset $D$ consisting of examples $x_i \in \{x_1, ..., x_{|D|}\}$. For each model $LLM_j$, we evaluate its performance by generating an output $o_i^j = LLM_j(x_i)$ for each example $x_i$. Each output $o_i^j$ has two associated quantities: the cost $c(o_i^j)$ of generating that output and the quality or performance $q(o_i^j)$ of the output itself. Through this process, we establish an expected cost $c_m$ and an expected quality $q_m$ for each model $LLM_m$ across the dataset $D$.
$$c_m = E[c(LLM_m(x)) | x \in D]$$
$$q_m = E[q(LLM_m(x)) | x \in D]$$
A \textit{router} $R$, define as a function, takes in a prompt $x$ and a set of parameters $\theta$, subsequently selecting the most suitable model $LLM_{i}$ from a set $L$ to complete the prompt, i.e. 
$$R_\theta(x) \mapsto LLM_{i} \in L.$$
The parameters $\theta$ typically include the maximum price the user is willing to pay, the desired latency, or a number of layers of neural networks for the router model, etc. More details of router parameters will be elaborated and discussed in Section \ref{router_design}.

The expected cost of a router $R_{\theta_1}$ across dataset $D$ is defined as 
$$c_{R_{\theta_1}}(D) = E[c(R_{\theta_1}(x)) | x \in D]$$ 
and the expected performance of a router $R_{\theta_1}$ can be defined similarly.

By experimenting with various router parameters $\theta_1,...,\theta_k$, we obtain a series of data points $(c_{R_{\theta_1}}, q_{R_{\theta_1}}),...,(c_{R_{\theta_k}}, q_{R_{\theta_k}})$ which can be graphically represented in the cost-quality ($c-q$) plane alongside the results of LLMs for comparative analysis.

\myparagraph{Linear Interpolation} The initial operation we introduce within this framework is \textit{linear interpolation}, which enables the computation of a weighted average between any two points on the cost-quality ($c-q$) plane.



\begin{figure*}[ht]
    \centering
    \includegraphics[width=\textwidth]{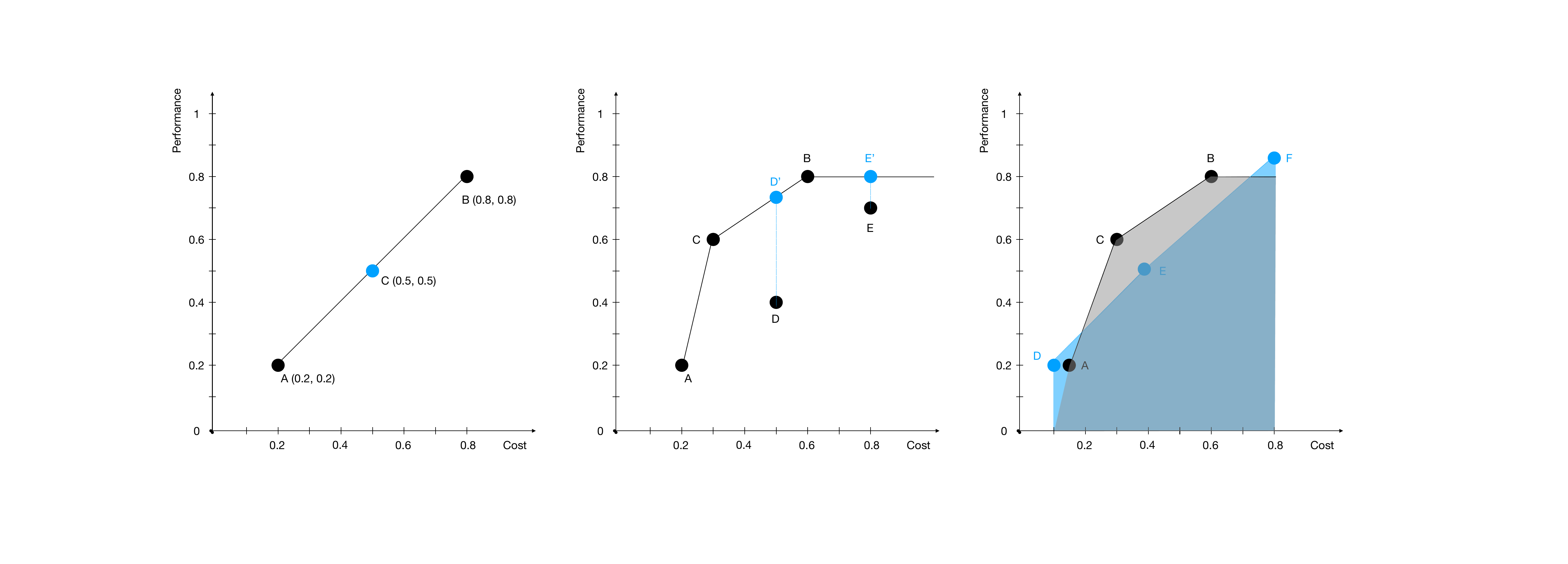}
    \vspace{-40pt}
    \caption{\textbf{Left}: linear interpolation is the process of achieving the cost-performance trade-off between any concrete routers. Point A and B are routers with different input parameters. To achieve the average of A and B, we build router C which routes to A or B with 50\% probability each, and it performs the average of A and B in expectation. \textbf{Middle}: Consider points A to E, we can construct the non-decreasing convex hull consisting of points A, B, and C. D and E as they can be replaced by a strictly superior affine combination of A, B, and C. \textbf{Right}: ABC and DEF are two routing systems (already convexified with ABC extrapolated to (0.1,0) for a fair comparison). To compare, we interpolate A and B to $c_{min}=0.1$ and $c_{max}=0.8$, respectively, and then calculate the area under the curve normalized by $c_{max}-c_{min}$ to derive AIQ.}
    \vspace{-10pt}
    \label{fig:ndch}
\end{figure*}
\vspace{5pt}

As illustrated by an example in the left of \fig{fig:ndch}, consider two routers, $R_{\theta_1}$ and $R_{\theta_2}$, we can formulate a third router, $R_{int}(R_{\theta_1}, R_{\theta_2})$, based on the following principle: given a prompt $x$ select $t \in [0, 1]$ such that:
$$
R_{int}(R_{\theta_1}, R_{\theta_2},t)(x) = \begin{cases} 
R_{\theta_1}(x), & \text{w.p. } t \\
R_{\theta_2}(x), & \text{w.p.  } 1-t 
\end{cases}
$$
Through the principle of linearity of expectation, we can deduce the expected cost of $R_{int}(R_{\theta_1}, R_{\theta_2},t)(x)$ in terms of $LLM_1$ and $LLM_2$: 
$$E[c_{R_{int}(x)} | x \in D] = t \cdot c_{R_{\theta_1}} + (1-t) \cdot  c_{R_{\theta_2}}$$
and the expected performance of $R_{int}(R_{\theta_1}, R_{\theta_2},t)(x)$ can be defined similarly.

Notably, for two data points $(c_1, q_1)$ and $(c_2, q_2)$ corresponding to $R_{\theta_1}$ and $R_{\theta_2}$ respectively, $R_{int}(t)$ can precisely interpolate any point along the line segment connecting $(c_1, q_1)$ and $(c_2, q_2)$. 

\myparagraph{Extrapolation} To ensure all routers can enrich our mathematical framework, we also introduce the \textit{extrapolation} operation, which enables all routers to extend to the cost domain $[0, \infty]$. For a given router $R_\theta$, we can trivially add more cost to the system without adding performance (for example repeat LLM generation $k$ times and only take the last generation as final output) and thus extend the cost to $\infty$. To extend the router to a smaller cost domain, we simply interpolate the null router (zero cost, zero performance) and $R_{\theta_1}$. Thus we are able to achieve any cost level between $[0, \infty]$ to when comparing routers with different domains.


It is essential to note that the routers discussed are functionally analogous to LLMs within this context, as both can be represented as coordinates in the cost-quality ($c-q$) plane.


\subsection{Non-Decreasing Convex Hull}


When working with multiple routers, it's feasible to construct any affine combination of points through linear interpolation among them. Specifically, for a set $S$ of points in the cost-quality ($c-q$) plane, these affine combinations can target any point $(c, q)$ in $\mathbb{R}^2$ lying within the convex hull formed by $S$. We identify $S_{ch} \subseteq S$ as the subset of points that delineate the vertices of this convex hull.



Furthermore, it's possible to configure a non-decreasing convex hull from $S_{ch}$, ensuring that for any two points $(c_1, q_1)$ and $(c_2, q_2)$ where $c_2 \geq c_1$, it follows that $q_2 \geq q_1$. Intuitively, if the extra cost of $c_2-c_1$ does not bring any performance improvement, it is advisable to simply extrapolate $(c_1, q_1)$ to the domain of $c_2$, and $(c_2, q_2)$ could be  $(c_2, q_1)$. An example is shown in the middle of \fig{fig:ndch}.

For a given routing system $R_1$, constituted by LLMs and routers plotted in the $c-q$ plane for dataset $D$, we can conceptualize a new routing system $\widetilde{R_1}$. This involves constructing routers $R_{\theta_1},...,R_{\theta_k}$, yielding points $(c_1,q_1),...,(c_k,q_k)$. By establishing a non-decreasing convex hull $S_{ndch}$ from these points and for any cost $c$ within the range $[c_{min}, c_{max}]$, optimal performance is attainable by interpolating between the two closest cost points. This process effectively creates a new routing system that spans the complete domain $[c_{min}, c_{max}]$. 



Given the framework established, we define the \textbf{Zero router}~($R_{zero}$) as a router that selects LLMs from $\{LLM_1, \ldots, LLM_m\}$ based on their collective non-decreasing convex hull. For a specified cost $c$, $R_{zero}$ provides a probabilistic mix of LLMs that maximizes expected output quality with a simple, mathematics-driven routing strategy. $R_{zero}$ serves as a basic benchmark for assessing the efficacy of other routing systems; a router is deemed significant only if it demonstrates superior performance compared to $R_{zero}$.

\subsection{Comparing Different Routing Systems}
Given the agnostic nature of our comparison framework towards the router's structure, routing systems can produce an assorted set of points on the $c-q$ plane that may be non-deterministic and non-parametric, complicating direct comparisons. Leveraging the methodologies delineated previously, we have the capacity to condense these disparate points into a streamlined function—specifically, a non-decreasing convex hull—and subsequently distill this representation into a singular metric that encapsulates the system's characteristics.

Routing systems often generate multiple points on the cost-quality ($c-q$) plane, making it difficult to compare the underlying systems. However, our framework allows us to transform these non-parametric points into a simpler function, specifically a non-decreasing convex hull, which can be characterized by a single numerical value.

Let's consider two different routing systems (for example KNN and MLP-based routers), $R_\theta$ where $\theta \in {\Theta}$, and $R_\lambda$ where $\lambda \in \Lambda$. To compare their effectiveness, we parametrize them by sampling from $\Theta, \Lambda$ to generate a set of points: $R_{\theta_1}, \dots, R_{\theta_k}$, and $R_{\lambda_1}, \dots, R_{\lambda_k}$. Then, we construct a non-decreasing convex hull for both groups, $\widetilde{R_\theta}$ and  $\widetilde{R_\lambda}$, defined on a shared domain $[c_{min}, c_{max}]$. 

 We define $AIQ$ (Average Improvement in Quality) for one of the routing systems as follows:
$$AIQ(R_\theta) = \frac{1}{c_{max} - c_{min}} \int^{c_{max}}_{c_{min}} \widetilde{R_\theta} \mathrm{ d}c$$
With the equation above, we can calculate AIQs for any group of routing systems to get a clear understanding of their relative performance, which is demonstrated in the right of \fig{fig:ndch}. Rather than performing complex graphic analysis, $AIQ$ allows users to measure router performance in a straightforward way.


\section{Benchmark Construction - \dataset}

To systematically assess router performance, we have developed a dataset, \dataset. This comprehensive dataset consists of a broad spectrum of tasks, including commonsense reasoning, knowledge-based language understanding, conversation, math, coding and retrieval-augmented generation (RAG). \dataset~is constructed by leveraging existing datasets that are widely recognized and utilized in the evaluation of leading LLMs, such as GPT-4, Gemini~\cite{gemini}, and Claude~\cite{claude}. This approach ensures that \dataset~is representative of the diverse challenges and requirements pertinent to mainstream LLM performance evaluation.

\subsection{Principles in benchmark construction}

The construction of \dataset~is guided by the following principles:
\begin{itemize}
    \setlength\itemsep{0em}
    \item Extensive Coverage: Our selection process identified a diverse array of fields where LLMs are widely utilized, aiming for wide-ranging applicability.
    \item Practical Relevance: The benchmarks chosen are of considerable significance to the industry's current applications of LLM systems, presenting a balanced challenge to the state-of-the-art LLMs, that is not too difficult nor too simplistic.
    \item Extensibility: \dataset~is designed for seamless integration of additional metrics, such as latencies and throughputs, ensuring adaptability to the evolving landscape of LLM.
\end{itemize} 


\subsection{Benchmark Dataset}
\label{benchmark_dataset}

For the initial release, we have curated a selection of 8 representative datasets from multiple different tasks. Detailed descriptions are in Appendix \ref{appendix:dataset_details}.
\begin{itemize}
    \setlength\itemsep{0em}
    \item \textbf{Commonsense Reasoning}: Hellaswag~\cite{hellaswag}, Winogrande~\cite{winogrande}, and ARC Challenge~\cite{arcc}
    \item \textbf {Knowledge-based Language Understanding}: MMLU~\cite{mmlu}
    \item \textbf{Conversation}: MT-Bench~\cite{mtbench}
    \item \textbf{Math}: GSM8K~\cite{gsm8k}
    \item \textbf{Coding}: MBPP~\cite{mbpp}
\end{itemize}

\textbf{RAG Dataset}: To evaluate routers in a more practical setting, we collected 800 user queries from one of Martian's clients, an LLM-assisted search company, and constructed an RAG dataset based on these queries. These queries cover topics including sports, history, media \& art, and politics, and all have ground truth answers. We then manually collected the ground truths, which were used to evaluate answers from selected groups of LLM and LLM-assisted search engines. This initiative is designed to assess the routers' performance in a complex "compound system" setting~\citep{compound-ai-blog} -- determining whether routers can adeptly navigate when retrieval abilities are also in play. For instance, when dealing with news published after the GPT-4 knowledge cutoff, routers are expected to more frequently opt for models that can access and search the latest internet-based information (e.g. sonar-medium-online).

\subsection{Dataset Construction Process}

For the compilation of our benchmark dataset, we perform inference with 14 different LLMs, with 3 of them specific to the RAG dataset\footnote{sonar-small-online and sonar-medium-online from Perplexity AI, You.com API}, 
including both open-source and proprietary models. This process was applied to each of the eight datasets and the RAG dataset enumerated in Section \ref{benchmark_dataset}, which is also illustrated in \fig{fig:teaser}. The selected LLMs are as follows and more details are in Appendix \ref{appendix:model_info}:

\myparagraph{Open Source Model:} Llama-70B-chat~\cite{touvron2023llama}, Mixtral-8x7B-chat~\cite{automix}, Yi-34B-chat~\cite{ai2024yi}, Code Llama-34B~\cite{rozière2023code}, Mistral-7B-chat~\cite{jiang2023mistral}, WizardLM-13B~\cite{xu2024wizardlm}

\myparagraph{Proprietary Model:} GPT-4, GPT-3.5-turbo~\cite{gpt4}, Claude-instant-v1, Claude-v1, Claude-v2~\cite{claude}, You.com API, sonar-small-online, sonar-medium-online.

In total, there are 405,467 samples in \dataset, covering 11 models, 8 datasets, and 64 tasks.

\begin{figure*}[h] 
    \begin{center}
      \begin{minipage}[c]{0.5\linewidth}
        \centering
        \includegraphics[width=\textwidth]{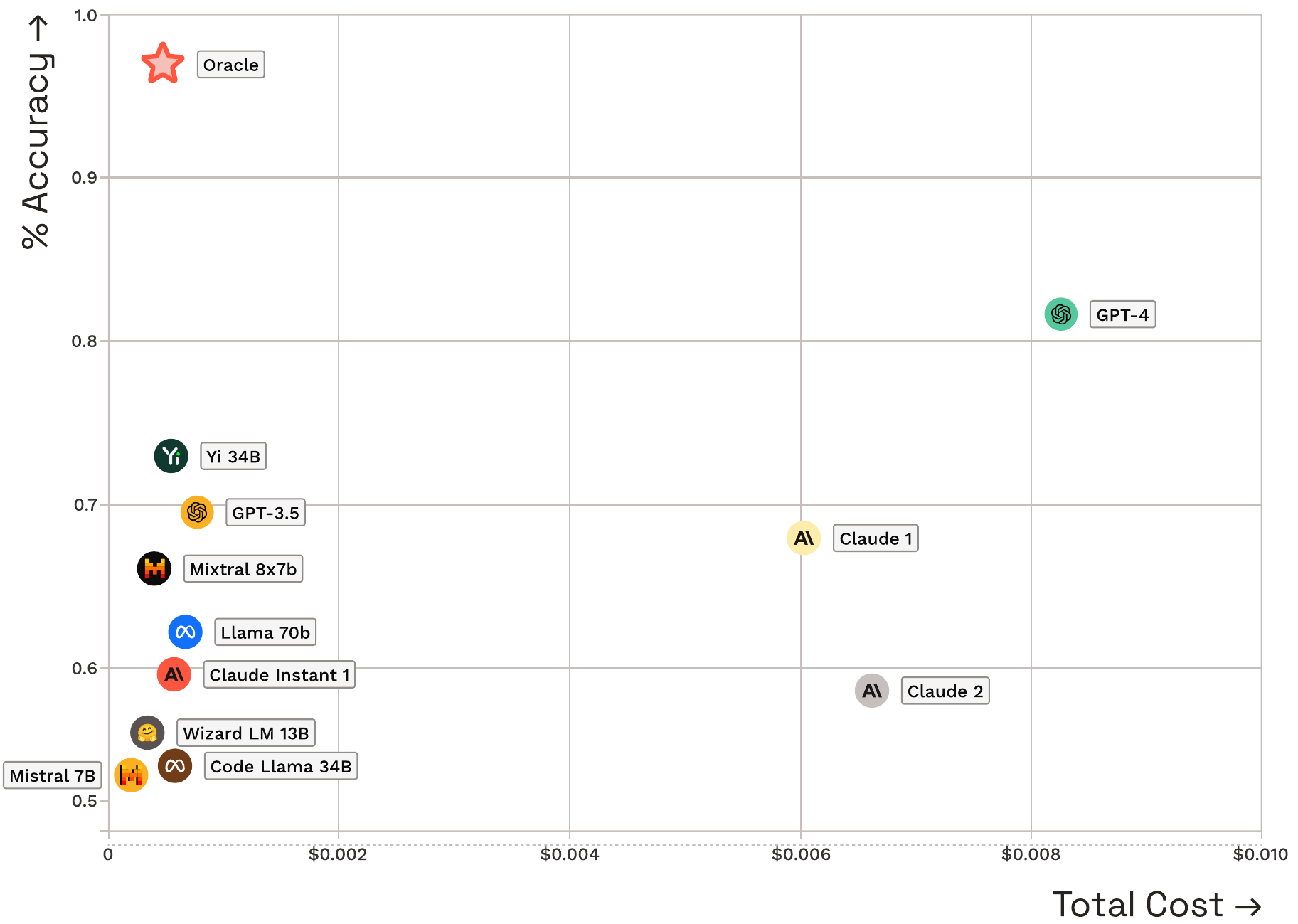}
        \label{fig:bedroom_qdiff}
      \end{minipage}\hfill
      \begin{minipage}[c]{0.5\linewidth}
        \centering
        \includegraphics[width=\textwidth]{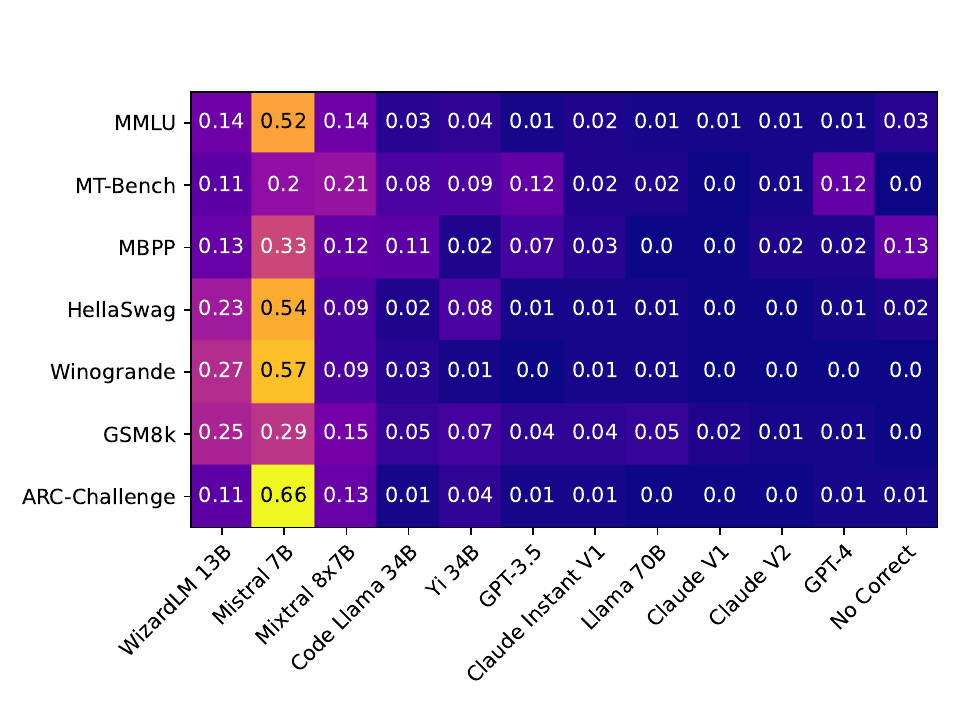}
        \label{fig:bedroom_lq}
      \end{minipage}
    \end{center}
    
    \vspace{-15pt}
    \caption{\textbf{Left}: Accuracy vs Total Cost of all the $11$ LLMs on \dataset. \textbf{Right}: The \textit{Oracle} LLMs selection frequency across the $7$ subsets in \dataset.}
    \vspace{-5pt}
    \label{fig:pilot}
\end{figure*}

\subsection{A Pilot Study: The Oracle Router}

We assessed the performance of various models across the eight datasets, with more details in ( \ref{appendix:evaluation_metric} and \ref{appendix:more_dataset}) while aggregate results are illustrated in \fig{fig:pilot}. The \textit{Oracle} represents the best possible router: the one that always routes to the best-performing LLM (if there are multiple of them, then route to the cheapest one).

\myparagraph{Result:} 
We note that the \textit{Oracle} router achieves near-optimal performance at a low cost, highlighting the potential for efficient routing among LLMs. Although proprietary models like GPT-4 offer superior performance, their higher cost than open-source alternatives is a significant drawback. Factors such as overalignment could also hurt the generation quality of proprietary models such as Claude 2 (refer to \append{appendix:refuse}). The heatmap in \fig{fig:pilot} illustrates that, despite WizardLM-13B and Mistral-7B achieving only about 50\% accuracy across tasks, their affordability leads to frequent selection by the \textit{Oracle}, prioritizing them when they provide correct responses. Moreover, the surprising observation that GPT-4 is seldom chosen suggests the existence of less expensive LLMs that can deliver high-quality answers for most queries. This underscores the substantial opportunity for enhancing LLM systems through cost-effective routing without sacrificing quality.

\section{Experiments}

\subsection{Predictive Router}
\label{router_design}
We propose a novel set of predictive routers that do not require the pre-generation of LLM outputs. Specifically, we introduce a router $R: x_i \to \text{LLM}$, constructed as follows: for an input $x_i$, the performance score for $LLM_j$ is calculated via:
\[
\text{performance score}_{ij} = \lambda \cdot P_{ij} - \text{cost}_j
\]
$P$ denotes the predicted performance of $LLM_j$ on sample $x_i$, with $\lambda$ representing the \textbf{willingness to pay (WTP)} parameter that delineates the cost-performance trade-off. A higher $\lambda$ indicates a preference for superior performance at a higher cost. We approximate the total cost using the cost per token metric. The routing decision for the predictive router is thus formulated as selecting the $LLM$ that optimizes the performance score.

To estimate $P$ for each input across models, we implemented two supervised regression approaches: \textbf{k-nearest neighbors (KNN)} and \textbf{multi-layer perceptron (MLP)} inspired by \cite{llmbenchmark}. We allocated a fraction of the dataset for training a performance predictor for each task, assessing its efficacy on the remainder.

Specifically, the \textbf{KNN router} estimates $\text{performance score}_{ij}$ by identifying the $k$ nearest samples in the training set $D_{train}$ and opting for $LLM_i$, demonstrating optimal performance within this subset.
$$
P_{\text{KNN}}(x_i) = \frac{1}{k} \sum_{x_j \in \text{NN}_k(x_i, D_{train})} q(o_j^i)
$$
Where $NN_k(x_i, D_{train})$ signifies the subset of $k$ nearest neighbors to the sample $x_i$ within the training dataset $D_{train}$.

\begin{figure*}[!ht]
\centering
\vspace{-5pt}
\includegraphics[width=1.0\linewidth]{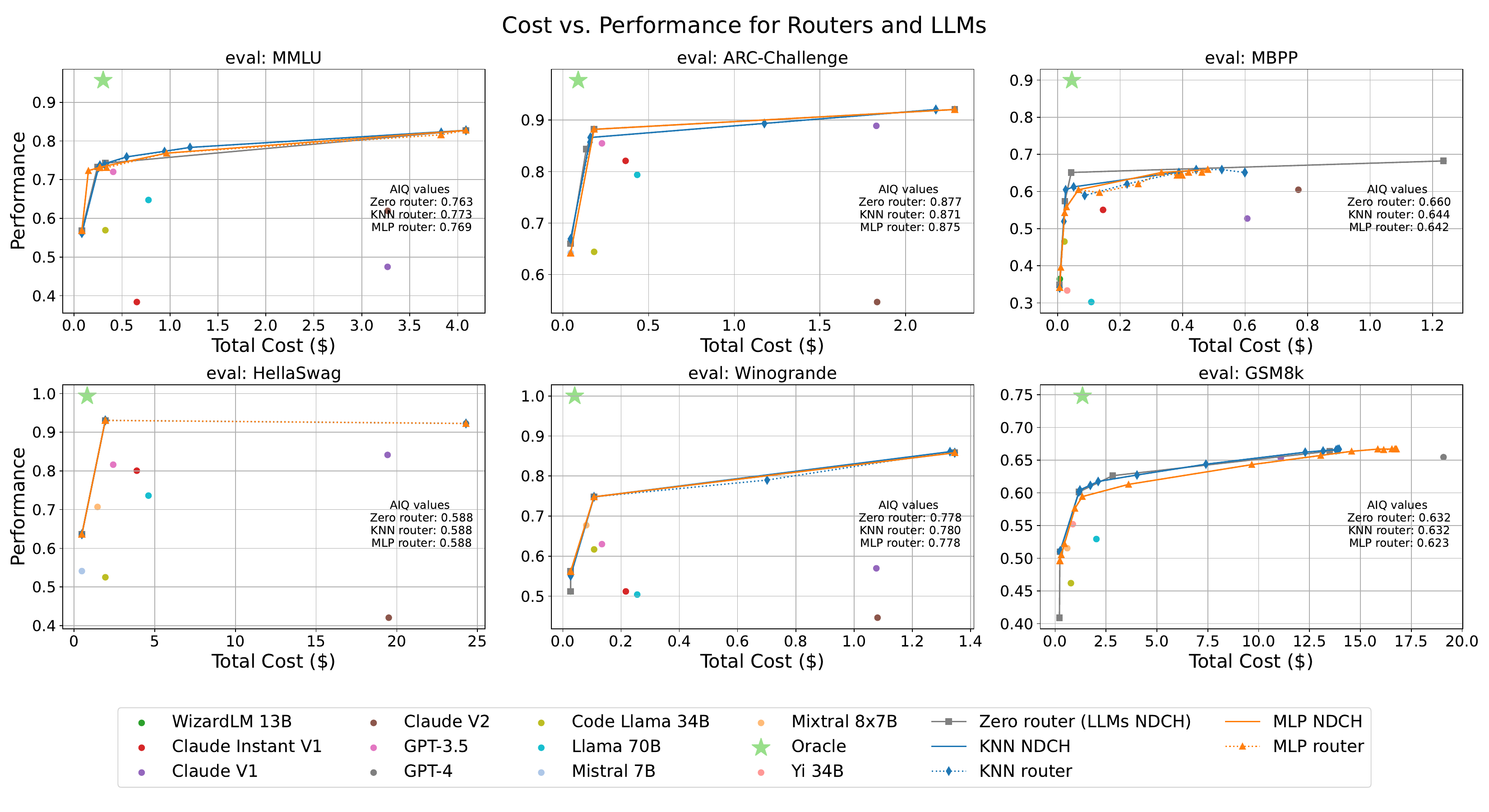}
\vspace{-25pt}
\caption{Total Cost vs. Performance for eleven models and KNN, MLP, and Zero routers on \dataset except for MT-Bench. For KNN and MLP, we tested different hyper-parameters, and the optimal results are displayed above. The AIQ values are calculated for all $3$ routers. NDCH stands for non-decreasing convex hull, represented by the solid lines. Dotted lines connect points with increasing willingness to pay.}
\label{results_main}
\vspace{-5pt}
\end{figure*}

Similarly, for \textbf{the MLP router}, we have trained a set of MLP models to predict the performance
$$
P_{\text{MLP}}(x_i) = f(W_n \cdot \sigma(... \cdot \sigma(W_1 \cdot x_i + b_1) ... + b_n)
$$
Those series of KNN and MLP routers are trained with varying hyperparameters, and we present the experimental results derived from the optimal hyperparameter configurations.

\subsection{Non-Predictive Routers} 
This category of routers generates answers from a sequence of Large Language Models (LLMs), evaluates these answers, and bases routing decisions on the evaluation outcomes. Drawing inspiration from \cite{frugalgpt, tabi}, we introduce a \textit{cascading router} comprising of a total cost parameter $T$, and a sequence of $m$ LLMs, denoted as $LLM_j: \text{text} \to \text{text}$, ranked from the least to the most expensive in terms of computational cost and expected accuracy. A key component of its operation is a scoring function $g: \text{text} \to [0,1]$ paired with a threshold $t$ (the "judge"). Upon receiving a request, it is initially processed by $LLM_1$. If g(o1)>t, the output o1 is selected, and the process terminates; otherwise, if the cumulative cost is still less than the total cost T, the router proceeds to the next LLM in the sequence and returns the current output if not. 



Although developing an effective scoring function $g$ for a specific task in a production setting presents challenges, within the context of this paper, the router possesses perfect knowledge of the final score, enabling it to consistently select the most cost-effective model that yields a satisfactory response (akin to an oracle). To simulate real-world performance more accurately, we introduce an error parameter $\epsilon \in [0,1]$. The adjusted scoring function $g_\epsilon(o)$ is defined as:
\[
g_\epsilon(o) = 
\begin{cases} 
1-g(o) & \text{with probability } \epsilon \\
g(o) & \text{with probability } 1-\epsilon 
\end{cases}
\]
A variant of the non-predictive router is overgenerate-and-rerank, which generates all potential outcomes from the LLM, assesses each, and outputs the optimal one as determined by a designated reward function. Although its practical application is limited due to significant costs, we will present its results for demonstration.

\subsection{Main Results}
\label{main_res}

\begin{figure*}[!ht]
    \centering
    \vspace{-5pt}
    \includegraphics[width=1.0\linewidth]{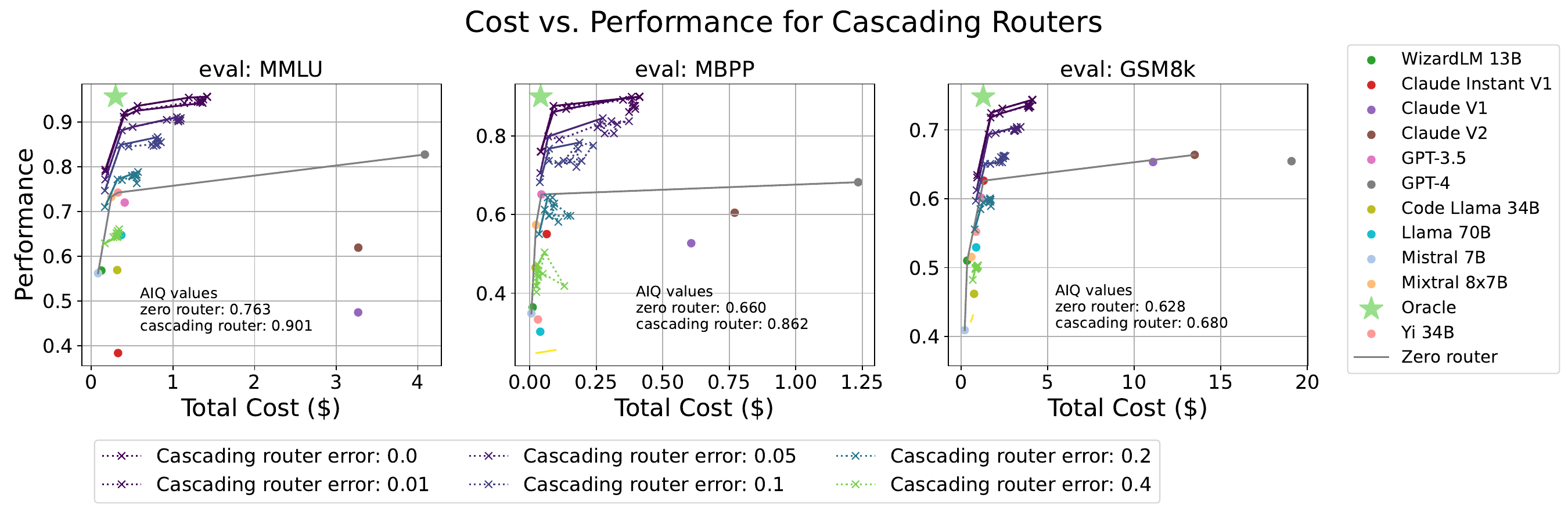}
    \vspace{-25pt}
    \caption{Total Cost vs Performance for eleven models and cascading routers on MMLU, MBPP, and GSM8K. Different error rates are tested, and the AIQ value is computed for Zero Router and zero error rate cascading router. The solid lines represent the non-decreasing convex hull and the dotted line represents points with increasing the maximum cost parameter.}
    \label{results_cas}
    \vspace{-5pt}
\end{figure*}

\myparagraph{Predictive Router}
With the KNN and MLP router design, we present the performances of predictive routers across all tasks (other than MT-Bench). The dataset for each task is randomly partitioned into two splits, where the routers are trained on $70\%$ and evaluated on the rest $30\%$. We exclude MT-Bench in this set of experiments due to its limited size in performing such a train-test partition. As shown in \fig{results_main}, both KNN routers and MLP routers achieve the level of performance to the best individual LLMs with lower or similar costs, demonstrating the effectiveness of the proposed routing solutions, despite their simplicity. However, none of the routing algorithms significantly outperform the baseline Zero router (The routers exhibit higher AIQ than the Zero router for MMLU and Winogrande, achieved comparable AIQ for Hellaswag and GSM8K, and underperform on ARC-Challenge and MBPP), the \textit{oracle} router consistently exceeds all other routers and LLMs in performance, underscoring the room for further advancements in routing algorithms design.


\myparagraph{Cascading Router}
We present results for cascading routers on MMLU, MBPP, and GSM8K in \fig{results_cas}. 
The results indicate that with each error rate, as the total cost \textit{T} increases, the cascading router's performance improves due to the availability of a larger budget for selecting more appropriate models. For lower error rates, the cascading router performs better than the Zero router, as evidenced by the higher AIQ value. The router with a zero error rate judge quickly approximates the performance of the \textit{Oracle} at the same cost and achieves comparable results as the cost further increases. \fig{results_cas} illustrates the cascading routers' effectiveness, showing they surpass both individual LLMs and the Zero router by a significant margin when the router's judge has an error rate of up to $0.1$. This indicates the routing technique's potential when paired with an effective judge.

However, as the judge's error rates increase, the performance of the cascading router may deteriorate rapidly, particularly when the error rate exceeds $0.2$. Achieving a sufficiently low error rate for certain real-world tasks to benefit from cascading routers might be challenging. Additionally, the sequence in which LLMs are chosen plays a crucial role in performance and offers room for optimization~\cite{frugalgpt}. Our findings present a simulated upper limit for this method, highlighting the potential and the necessity of exploring the optimal implementation of cascading routers for specific applications.



\subsection{RAG Results}
\label{main_rag_result}
\begin{figure}[h] 
    
    \centering
    \includegraphics[width=0.5\textwidth]{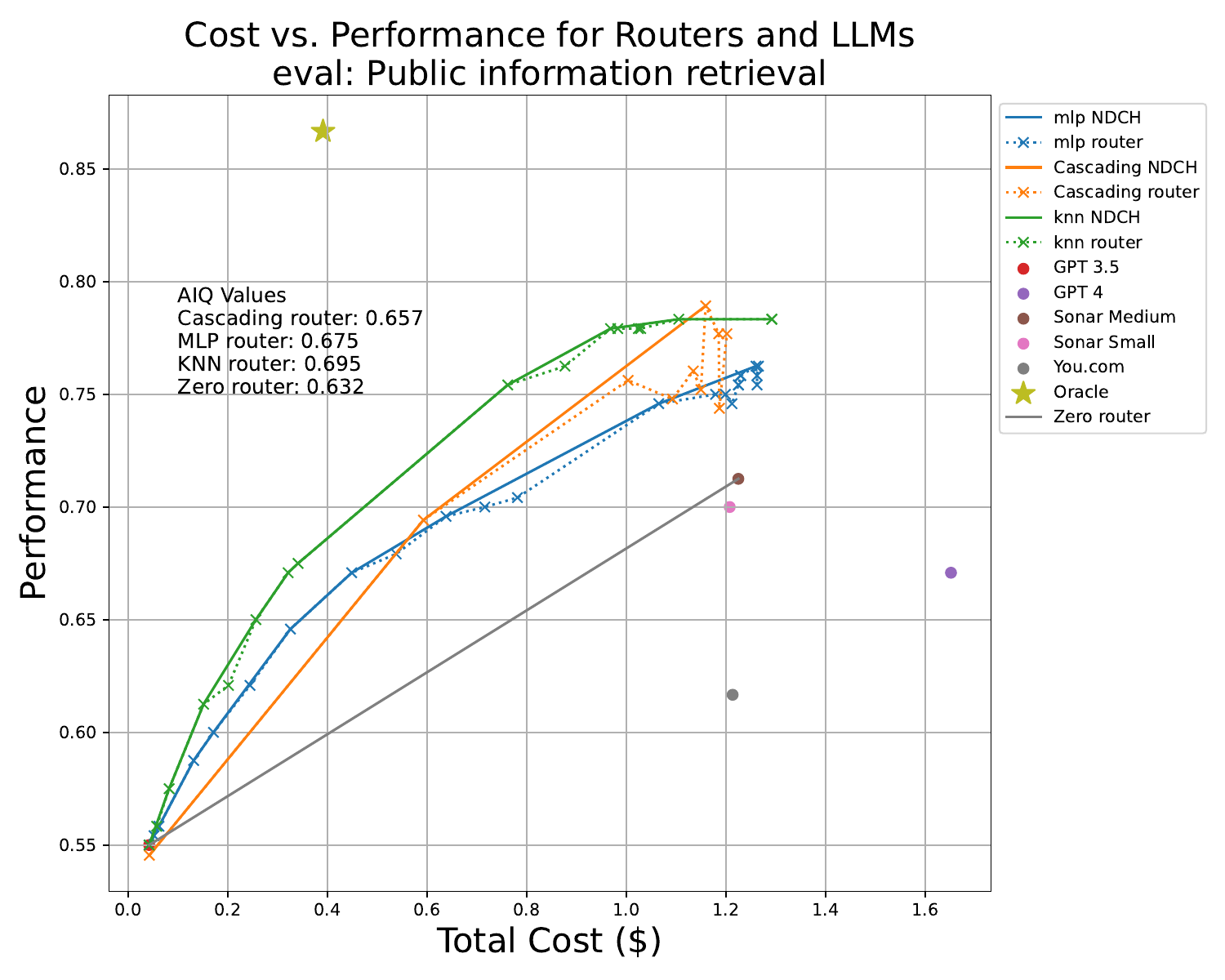}
    \vspace{-15pt}
    \caption{Total Cost vs Performance for five models and four routers on the RAG dataset. The AIQ values are also calculated for all four routers. NDCH represents the non-decreasing convex hull. 
    }
    \vspace{-10pt}
    \label{results_rag}
\end{figure}

Building on the results above, we simultaneously compared various router types, including predictive and cascading routers, on the RAG dataset. We used the same setting for KNN and MLP routers while selecting an error rate 0.2 for cascading routers. We randomly partitioned the RAG dataset into two splits: 70\% for training predictive routers and 30\% for evaluating all routers. \fig{results_rag} demonstrates that all routers significantly improve compared to the Zero Router. Further analysis shows that the routers can identify time-sensitive features (like "2024") in user queries and route to online models for time-sensitive queries and GPT-4/GPT-3.5 for time-insensitive queries. Our findings highlight the potential of model routing to enhance LLM applications within the "Compound AI Systems"~\cite{compound-ai-blog} scenario.




\section{Limitations and Future Work}

\dataset~currently only focuses on performance and economic cost. It is meaningful to include more evaluation criteria, such as latency, throughput, and others, to capture a more comprehensive understanding of router capabilities and limitations. There are also many LLMs and tasks that are not included in  \dataset~due to the limitation of time, and future iterations of this benchmark would include datasets that cover more tasks to evaluate the ever-growing capability of LLMs effectively and also to add newer LLMs as they are being released.

Our current work only evaluates the efficacy of predictive and cascading routers, yet considerable scope exists for investigating further router designs, as highlighted in \sect{main_res}. Delving into more advanced router designs is crucial for enhancing routing efficiency. Notably, our evaluation within the RAG context was limited to models possessing inherent retrieval capabilities. Addressing the challenge of implementing two-stage routing, which encompasses retrievers and LLMs, remains critical. This approach could significantly refine router evaluations on standard RAG tasks, including HotpotQA~\cite{yang2018hotpotqa} and NaturalQuestions~\cite{kwiatkowski2019natural}, by ensuring more precise assessments.

Furthermore, although the seven datasets in~\dataset  offer broad coverage across various tasks, incorporating domain-specific tasks that require long-tail skills, like translation of low-resource languages, could reveal additional intriguing aspects of LLM routing. This enhancement would align the benchmark more closely with real-world application scenarios. Efforts to integrate such tasks in future versions are planned. 


\section{Conclusion}
We present \dataset, a benchmark specifically designed to evaluate routers for multi-LLM systems. By addressing the critical need for standardized evaluation in this domain, our benchmark provides a comprehensive dataset and a theoretical framework designed for the nuanced analysis of router cost-efficiency and performance. The insights from our study shed light on the effectiveness of various routing strategies and revealed promising early results in some tasks. This work establishes a robust and scalable benchmark for router evaluation and aims to facilitate future progress in the efficient and cost-effective deployment of Large Language Models.


\bibliography{reference}

\begin{thebibliography}{52}
\providecommand{\natexlab}[1]{#1}
\providecommand{\url}[1]{\texttt{#1}}
\expandafter\ifx\csname urlstyle\endcsname\relax
  \providecommand{\doi}[1]{doi: #1}\else
  \providecommand{\doi}{doi: \begingroup \urlstyle{rm}\Url}\fi

\bibitem[AI et~al.(2024)AI, :, Young, Chen, Li, Huang, Zhang, Zhang, Li, Zhu, Chen, Chang, Yu, Liu, Liu, Yue, Yang, Yang, Yu, Xie, Huang, Hu, Ren, Niu, Nie, Xu, Liu, Wang, Cai, Gu, Liu, and Dai]{ai2024yi}
01. AI, :, Alex Young, Bei Chen, Chao Li, Chengen Huang, Ge~Zhang, Guanwei Zhang, Heng Li, Jiangcheng Zhu, Jianqun Chen, Jing Chang, Kaidong Yu, Peng Liu, Qiang Liu, Shawn Yue, Senbin Yang, Shiming Yang, Tao Yu, Wen Xie, Wenhao Huang, Xiaohui Hu, Xiaoyi Ren, Xinyao Niu, Pengcheng Nie, Yuchi Xu, Yudong Liu, Yue Wang, Yuxuan Cai, Zhenyu Gu, Zhiyuan Liu, and Zonghong Dai.
\newblock Yi: Open foundation models by 01.ai.
\newblock \emph{arXiv preprint arXiv: 2403.04652}, 2024.

\bibitem[Alzahrani et~al.(2024)Alzahrani, Alyahya, Alnumay, Alrashed, Alsubaie, Almushaykeh, Mirza, Alotaibi, Altwairesh, Alowisheq, Bari, and Khan]{alzahrani2024benchmarks}
Norah Alzahrani, Hisham~Abdullah Alyahya, Yazeed Alnumay, Sultan Alrashed, Shaykhah Alsubaie, Yusef Almushaykeh, Faisal Mirza, Nouf Alotaibi, Nora Altwairesh, Areeb Alowisheq, M~Saiful Bari, and Haidar Khan.
\newblock When benchmarks are targets: Revealing the sensitivity of large language model leaderboards.
\newblock \emph{arXiv preprint arXiv: 2402.01781}, 2024.

\bibitem[Anthropic(2023)]{claude}
Anthropic.
\newblock Model card and evaluations for claude models, 2023.
\newblock URL \url{https://www-cdn.anthropic.com/files/4zrzovbb/website/bd2a28d2535bfb0494cc8e2a3bf135d2e7523226.pdf}.

\bibitem[Austin et~al.(2021)Austin, Odena, Nye, Bosma, Michalewski, Dohan, Jiang, Cai, Terry, Le, and Sutton]{mbpp}
Jacob Austin, Augustus Odena, Maxwell Nye, Maarten Bosma, Henryk Michalewski, David Dohan, Ellen Jiang, Carrie Cai, Michael Terry, Quoc Le, and Charles Sutton.
\newblock Program synthesis with large language models.
\newblock 2021.

\bibitem[Bubeck et~al.(2023)Bubeck, Chandrasekaran, Eldan, Gehrke, Horvitz, Kamar, Lee, Lee, Li, Lundberg, et~al.]{bubeck2023sparks}
S{\'e}bastien Bubeck, Varun Chandrasekaran, Ronen Eldan, Johannes Gehrke, Eric Horvitz, Ece Kamar, Peter Lee, Yin~Tat Lee, Yuanzhi Li, Scott Lundberg, et~al.
\newblock Sparks of artificial general intelligence: Early experiments with gpt-4.
\newblock \emph{arXiv preprint arXiv:2303.12712}, 2023.

\bibitem[Chen et~al.(2023)Chen, Zaharia, and Zou]{frugalgpt}
Lingjiao Chen, Matei Zaharia, and James Zou.
\newblock Frugalgpt: How to use large language models while reducing cost and improving performance.
\newblock \emph{arXiv preprint arXiv: Arxiv-2305.05176}, 2023.

\bibitem[Clark et~al.(2018)Clark, Cowhey, Etzioni, Khot, Sabharwal, Schoenick, and Tafjord]{arcc}
Peter Clark, Isaac Cowhey, Oren Etzioni, Tushar Khot, Ashish Sabharwal, Carissa Schoenick, and Oyvind Tafjord.
\newblock Think you have solved question answering? try arc, the ai2 reasoning challenge.
\newblock \emph{arXiv preprint arXiv:1803.05457}, 2018.

\bibitem[Cobbe et~al.(2021)Cobbe, Kosaraju, Bavarian, Chen, Jun, Kaiser, Plappert, Tworek, Hilton, Nakano, Hesse, and Schulman]{gsm8k}
Karl Cobbe, Vineet Kosaraju, Mohammad Bavarian, Mark Chen, Heewoo Jun, Lukasz Kaiser, Matthias Plappert, Jerry Tworek, Jacob Hilton, Reiichiro Nakano, Christopher Hesse, and John Schulman.
\newblock Training verifiers to solve math word problems.
\newblock 2021.

\bibitem[Du et~al.(2022)Du, Huang, Dai, Tong, Lepikhin, Xu, Krikun, Zhou, Yu, Firat, Zoph, Fedus, Bosma, Zhou, Wang, Wang, Webster, Pellat, Robinson, Meier-Hellstern, Duke, Dixon, Zhang, Le, Wu, Chen, and Cui]{glam}
Nan Du, Yanping Huang, Andrew~M. Dai, Simon Tong, Dmitry Lepikhin, Yuanzhong Xu, Maxim Krikun, Yanqi Zhou, Adams~Wei Yu, Orhan Firat, Barret Zoph, Liam Fedus, Maarten Bosma, Zongwei Zhou, Tao Wang, Yu~Emma Wang, Kellie Webster, Marie Pellat, Kevin Robinson, Kathleen Meier-Hellstern, Toju Duke, Lucas Dixon, Kun Zhang, Quoc~V Le, Yonghui Wu, Zhifeng Chen, and Claire Cui.
\newblock Glam: Efficient scaling of language models with mixture-of-experts.
\newblock 2022.

\bibitem[Eigen et~al.(2014)Eigen, Ranzato, and Sutskever]{dMoE}
David Eigen, Marc'Aurelio Ranzato, and Ilya Sutskever.
\newblock Learning factored representations in a deep mixture of experts.
\newblock 2014.

\bibitem[Fedus et~al.(2022)Fedus, Zoph, and Shazeer]{switch}
William Fedus, Barret Zoph, and Noam Shazeer.
\newblock Switch transformers: Scaling to trillion parameter models with simple and efficient sparsity.
\newblock 2022.

\bibitem[Hari \& Thomson(2023)Hari and Thomson]{tryage}
Surya~Narayanan Hari and Matt Thomson.
\newblock Tryage: Real-time, intelligent routing of user prompts to large language models.
\newblock 2023.

\bibitem[Hendrycks et~al.(2021)Hendrycks, Burns, Basart, Zou, Mazeika, Song, and Steinhardt]{mmlu}
Dan Hendrycks, Collin Burns, Steven Basart, Andy Zou, Mantas Mazeika, Dawn Song, and Jacob Steinhardt.
\newblock Measuring massive multitask language understanding.
\newblock In \emph{9th International Conference on Learning Representations}, 2021.
\newblock URL \url{https://openreview.net/forum?id=d7KBjmI3GmQ}.

\bibitem[Jiang et~al.(2023{\natexlab{a}})Jiang, Sablayrolles, Mensch, Bamford, Chaplot, de~las Casas, Bressand, Lengyel, Lample, Saulnier, Lavaud, Lachaux, Stock, Scao, Lavril, Wang, Lacroix, and Sayed]{jiang2023mistral}
Albert~Q. Jiang, Alexandre Sablayrolles, Arthur Mensch, Chris Bamford, Devendra~Singh Chaplot, Diego de~las Casas, Florian Bressand, Gianna Lengyel, Guillaume Lample, Lucile Saulnier, Lélio~Renard Lavaud, Marie-Anne Lachaux, Pierre Stock, Teven~Le Scao, Thibaut Lavril, Thomas Wang, Timothée Lacroix, and William~El Sayed.
\newblock Mistral 7b.
\newblock \emph{arXiv preprint arXiv: 2310.06825}, 2023{\natexlab{a}}.

\bibitem[Jiang et~al.(2023{\natexlab{b}})Jiang, Ren, and Lin]{llmblender}
Dongfu Jiang, Xiang Ren, and Bill~Yuchen Lin.
\newblock {LLM}-blender: Ensembling large language models with pairwise ranking and generative fusion.
\newblock In Anna Rogers, Jordan Boyd-Graber, and Naoaki Okazaki (eds.), \emph{Proceedings of the 61st Annual Meeting of the Association for Computational Linguistics (Volume 1: Long Papers)}, pp.\  14165--14178, Toronto, Canada, July 2023{\natexlab{b}}. Association for Computational Linguistics.
\newblock \doi{10.18653/v1/2023.acl-long.792}.
\newblock URL \url{https://aclanthology.org/2023.acl-long.792}.

\bibitem[Khattab et~al.(2024)Khattab, Singhvi, Maheshwari, Zhang, Santhanam, A, Haq, Sharma, Joshi, Moazam, Miller, Zaharia, and Potts]{khattab2024dspy}
Omar Khattab, Arnav Singhvi, Paridhi Maheshwari, Zhiyuan Zhang, Keshav Santhanam, Sri~Vardhamanan A, Saiful Haq, Ashutosh Sharma, Thomas~T. Joshi, Hanna Moazam, Heather Miller, Matei Zaharia, and Christopher Potts.
\newblock {DSP}y: Compiling declarative language model calls into state-of-the-art pipelines.
\newblock In \emph{The Twelfth International Conference on Learning Representations}, 2024.
\newblock URL \url{https://openreview.net/forum?id=sY5N0zY5Od}.

\bibitem[Kim et~al.(2023)Kim, Hooper, Gholami, Dong, Li, Shen, Mahoney, and Keutzer]{kim2023squeezellm}
Sehoon Kim, Coleman Hooper, Amir Gholami, Zhen Dong, Xiuyu Li, Sheng Shen, Michael~W. Mahoney, and Kurt Keutzer.
\newblock Squeezellm: Dense-and-sparse quantization.
\newblock \emph{arXiv preprint arXiv: 2306.07629}, 2023.

\bibitem[Kwiatkowski et~al.(2019)Kwiatkowski, Palomaki, Redfield, Collins, Parikh, Alberti, Epstein, Polosukhin, Devlin, Lee, Toutanova, Jones, Kelcey, Chang, Dai, Uszkoreit, Le, and Petrov]{kwiatkowski2019natural}
Tom Kwiatkowski, Jennimaria Palomaki, Olivia Redfield, Michael Collins, Ankur Parikh, Chris Alberti, Danielle Epstein, Illia Polosukhin, Jacob Devlin, Kenton Lee, Kristina Toutanova, Llion Jones, Matthew Kelcey, Ming-Wei Chang, Andrew~M. Dai, Jakob Uszkoreit, Quoc Le, and Slav Petrov.
\newblock Natural questions: A benchmark for question answering research.
\newblock \emph{Transactions of the Association for Computational Linguistics}, 7:\penalty0 452--466, 2019.
\newblock \doi{10.1162/tacl_a_00276}.
\newblock URL \url{https://aclanthology.org/Q19-1026}.

\bibitem[Kwon et~al.(2023)Kwon, Li, Zhuang, Sheng, Zheng, Yu, Gonzalez, Zhang, and Stoica]{kwon2023efficient}
Woosuk Kwon, Zhuohan Li, Siyuan Zhuang, Ying Sheng, Lianmin Zheng, Cody~Hao Yu, Joseph Gonzalez, Hao Zhang, and Ion Stoica.
\newblock Efficient memory management for large language model serving with pagedattention.
\newblock In \emph{Proceedings of the 29th Symposium on Operating Systems Principles}, SOSP '23, pp.\  611–626, 2023.

\bibitem[Lee et~al.(2023)Lee, Cheng, and Ostendorf]{orchestrallm}
Chia-Hsuan Lee, Hao Cheng, and Mari Ostendorf.
\newblock Orchestrallm: Efficient orchestration of language models for dialogue state tracking.
\newblock 2023.

\bibitem[Levesque et~al.(2012)Levesque, Davis, and Morgenstern]{winograd}
Hector Levesque, Ernest Davis, and Leora Morgenstern.
\newblock The winograd schema challenge.
\newblock In \emph{Thirteenth international conference on the principles of knowledge representation and reasoning}, 2012.

\bibitem[Lin et~al.(2023)Lin, Tang, Tang, Yang, Dang, and Han]{lin2023awq}
Ji~Lin, Jiaming Tang, Haotian Tang, Shang Yang, Xingyu Dang, and Song Han.
\newblock Awq: Activation-aware weight quantization for llm compression and acceleration.
\newblock \emph{arXiv preprint arXiv: 2306.00978}, 2023.

\bibitem[Lu et~al.(2023)Lu, Yuan, Lin, Lin, Yuan, Zhou, and Zhou]{routingtoexpert}
Keming Lu, Hongyi Yuan, Runji Lin, Junyang Lin, Zheng Yuan, Chang Zhou, and Jingren Zhou.
\newblock Routing to the expert: Efficient reward-guided ensemble of large language models.
\newblock 2023.

\bibitem[Lu et~al.(2024)Lu, Liu, Liusie, Raina, Mudupalli, Zhang, and Beauchamp]{blending}
Xiaoding Lu, Zongyi Liu, Adian Liusie, Vyas Raina, Vineet Mudupalli, Yuwen Zhang, and William Beauchamp.
\newblock Blending is all you need: Cheaper, better alternative to trillion-parameters llm.
\newblock \emph{arXiv preprint arXiv: 2401.02994}, 2024.

\bibitem[Madaan et~al.(2023)Madaan, Aggarwal, Anand, Potharaju, Mishra, Zhou, Gupta, Rajagopal, Kappaganthu, Yang, Upadhyay, Mausam, and Faruqui]{automix}
Aman Madaan, Pranjal Aggarwal, Ankit Anand, Srividya~Pranavi Potharaju, Swaroop Mishra, Pei Zhou, Aditya Gupta, Dheeraj Rajagopal, Karthik Kappaganthu, Yiming Yang, Shyam Upadhyay, Mausam, and Manaal Faruqui.
\newblock Automix: Automatically mixing language models.
\newblock 2023.

\bibitem[OpenAI(2023)]{gpt4}
OpenAI.
\newblock {GPT-4} technical report.
\newblock \emph{CoRR}, abs/2303.08774, 2023.
\newblock \doi{10.48550/ARXIV.2303.08774}.
\newblock URL \url{https://doi.org/10.48550/arXiv.2303.08774}.

\bibitem[Rafailov et~al.(2023)Rafailov, Sharma, Mitchell, Ermon, Manning, and Finn]{rafailov2023direct}
Rafael Rafailov, Archit Sharma, Eric Mitchell, Stefano Ermon, Christopher~D. Manning, and Chelsea Finn.
\newblock Direct preference optimization: Your language model is secretly a reward model.
\newblock 2023.

\bibitem[Reimers \& Gurevych(2019)Reimers and Gurevych]{reimers-2019-sentence-bert}
Nils Reimers and Iryna Gurevych.
\newblock Sentence-bert: Sentence embeddings using siamese bert-networks.
\newblock In \emph{Proceedings of the 2019 Conference on Empirical Methods in Natural Language Processing}. Association for Computational Linguistics, 11 2019.
\newblock URL \url{https://arxiv.org/abs/1908.10084}.

\bibitem[Rozière et~al.(2023)Rozière, Gehring, Gloeckle, Sootla, Gat, Tan, Adi, Liu, Sauvestre, Remez, Rapin, Kozhevnikov, Evtimov, Bitton, Bhatt, Ferrer, Grattafiori, Xiong, Défossez, Copet, Azhar, Touvron, Martin, Usunier, Scialom, and Synnaeve]{rozière2023code}
Baptiste Rozière, Jonas Gehring, Fabian Gloeckle, Sten Sootla, Itai Gat, Xiaoqing~Ellen Tan, Yossi Adi, Jingyu Liu, Romain Sauvestre, Tal Remez, Jérémy Rapin, Artyom Kozhevnikov, Ivan Evtimov, Joanna Bitton, Manish Bhatt, Cristian~Canton Ferrer, Aaron Grattafiori, Wenhan Xiong, Alexandre Défossez, Jade Copet, Faisal Azhar, Hugo Touvron, Louis Martin, Nicolas Usunier, Thomas Scialom, and Gabriel Synnaeve.
\newblock Code llama: Open foundation models for code.
\newblock \emph{arXiv preprint arXiv: 2308.12950}, 2023.

\bibitem[Sakaguchi et~al.(2021)Sakaguchi, Bras, Bhagavatula, and Choi]{winogrande}
Keisuke Sakaguchi, Ronan~Le Bras, Chandra Bhagavatula, and Yejin Choi.
\newblock Winogrande: An adversarial winograd schema challenge at scale.
\newblock \emph{Communications of the ACM}, 64\penalty0 (9):\penalty0 99--106, 2021.

\bibitem[{\v{S}}akota et~al.(2023){\v{S}}akota, Peyrard, and West]{flyswat}
Marija {\v{S}}akota, Maxime Peyrard, and Robert West.
\newblock Fly-swat or cannon? cost-effective language model choice via meta-modeling.
\newblock \emph{arXiv preprint arXiv:2308.06077}, 2023.

\bibitem[Shazeer et~al.(2017)Shazeer, Mirhoseini, Maziarz, Davis, Le, Hinton, and Dean]{outrageously}
Noam Shazeer, Azalia Mirhoseini, Krzysztof Maziarz, Andy Davis, Quoc~V. Le, Geoffrey~E. Hinton, and Jeff Dean.
\newblock Outrageously large neural networks: The sparsely-gated mixture-of-experts layer.
\newblock In \emph{5th International Conference on Learning Representations, {ICLR} 2017, Toulon, France, April 24-26, 2017, Conference Track Proceedings}. OpenReview.net, 2017.
\newblock URL \url{https://openreview.net/forum?id=B1ckMDqlg}.

\bibitem[Shen et~al.(2023)Shen, Hou, Zhou, Du, Longpre, Wei, Chung, Zoph, Fedus, Chen, Vu, Wu, Chen, Webson, Li, Zhao, Yu, Keutzer, Darrell, and Zhou]{shenmoeit}
Sheng Shen, Le~Hou, Yanqi Zhou, Nan Du, Shayne Longpre, Jason Wei, Hyung~Won Chung, Barret Zoph, William Fedus, Xinyun Chen, Tu~Vu, Yuexin Wu, Wuyang Chen, Albert Webson, Yunxuan Li, Vincent Zhao, Hongkun Yu, Kurt Keutzer, Trevor Darrell, and Denny Zhou.
\newblock Mixture-of-experts meets instruction tuning: A winning combination for large language models.
\newblock 2023.

\bibitem[Shnitzer et~al.(2023)Shnitzer, Ou, Silva, Soule, Sun, Solomon, Thompson, and Yurochkin]{llmbenchmark}
Tal Shnitzer, Anthony Ou, Mírian Silva, Kate Soule, Yuekai Sun, Justin Solomon, Neil Thompson, and Mikhail Yurochkin.
\newblock Large language model routing with benchmark datasets.
\newblock 2023.

\bibitem[Si et~al.(2023)Si, Shi, Zhao, Zettlemoyer, and Boyd-Graber]{si2023getting}
Chenglei Si, Weijia Shi, Chen Zhao, Luke Zettlemoyer, and Jordan~L. Boyd-Graber.
\newblock Getting more out of mixture of language model reasoning experts.
\newblock \emph{Conference on Empirical Methods in Natural Language Processing}, 2023.
\newblock \doi{10.18653/v1/2023.findings-emnlp.552}.

\bibitem[Singhvi et~al.(2023)Singhvi, Shetty, Tan, Potts, Sen, Zaharia, and Khattab]{singhvi2023dspy}
Arnav Singhvi, Manish Shetty, Shangyin Tan, Christopher Potts, Koushik Sen, Matei Zaharia, and Omar Khattab.
\newblock Dspy assertions: Computational constraints for self-refining language model pipelines.
\newblock \emph{arXiv preprint arXiv: 2312.13382}, 2023.

\bibitem[Taori et~al.(2023)Taori, Gulrajani, Zhang, Dubois, Li, Guestrin, Liang, and Hashimoto]{taori2023stanford}
Rohan Taori, Ishaan Gulrajani, Tianyi Zhang, Yann Dubois, Xuechen Li, Carlos Guestrin, Percy Liang, and Tatsunori~B Hashimoto.
\newblock Stanford alpaca: An instruction-following llama model.
\newblock 2023.

\bibitem[Team et~al.(2023)Team, Anil, Borgeaud, Wu, Alayrac, Yu, Soricut, Schalkwyk, Dai, Hauth, et~al.]{gemini}
Gemini Team, Rohan Anil, Sebastian Borgeaud, Yonghui Wu, Jean-Baptiste Alayrac, Jiahui Yu, Radu Soricut, Johan Schalkwyk, Andrew~M Dai, Anja Hauth, et~al.
\newblock Gemini: a family of highly capable multimodal models.
\newblock \emph{arXiv preprint arXiv:2312.11805}, 2023.

\bibitem[Touvron et~al.(2023)Touvron, Martin, Stone, Albert, Almahairi, Babaei, Bashlykov, Batra, Bhargava, Bhosale, Bikel, Blecher, Ferrer, Chen, Cucurull, Esiobu, Fernandes, Fu, Fu, Fuller, Gao, Goswami, Goyal, Hartshorn, Hosseini, Hou, Inan, Kardas, Kerkez, Khabsa, Kloumann, Korenev, Koura, Lachaux, Lavril, Lee, Liskovich, Lu, Mao, Martinet, Mihaylov, Mishra, Molybog, Nie, Poulton, Reizenstein, Rungta, Saladi, Schelten, Silva, Smith, Subramanian, Tan, Tang, Taylor, Williams, Kuan, Xu, Yan, Zarov, Zhang, Fan, Kambadur, Narang, Rodriguez, Stojnic, Edunov, and Scialom]{touvron2023llama}
Hugo Touvron, Louis Martin, Kevin Stone, Peter Albert, Amjad Almahairi, Yasmine Babaei, Nikolay Bashlykov, Soumya Batra, Prajjwal Bhargava, Shruti Bhosale, Dan Bikel, Lukas Blecher, Cristian~Canton Ferrer, Moya Chen, Guillem Cucurull, David Esiobu, Jude Fernandes, Jeremy Fu, Wenyin Fu, Brian Fuller, Cynthia Gao, Vedanuj Goswami, Naman Goyal, Anthony Hartshorn, Saghar Hosseini, Rui Hou, Hakan Inan, Marcin Kardas, Viktor Kerkez, Madian Khabsa, Isabel Kloumann, Artem Korenev, Punit~Singh Koura, Marie-Anne Lachaux, Thibaut Lavril, Jenya Lee, Diana Liskovich, Yinghai Lu, Yuning Mao, Xavier Martinet, Todor Mihaylov, Pushkar Mishra, Igor Molybog, Yixin Nie, Andrew Poulton, Jeremy Reizenstein, Rashi Rungta, Kalyan Saladi, Alan Schelten, Ruan Silva, Eric~Michael Smith, Ranjan Subramanian, Xiaoqing~Ellen Tan, Binh Tang, Ross Taylor, Adina Williams, Jian~Xiang Kuan, Puxin Xu, Zheng Yan, Iliyan Zarov, Yuchen Zhang, Angela Fan, Melanie Kambadur, Sharan Narang, Aurelien Rodriguez, Robert Stojnic, Sergey Edunov, and Thomas
  Scialom.
\newblock Llama 2: Open foundation and fine-tuned chat models.
\newblock 2023.

\bibitem[Wang et~al.(2022)Wang, Wei, Schuurmans, Le, Chi, and Zhou]{wang2023selfconsistency}
Xuezhi Wang, Jason Wei, D.~Schuurmans, Quoc Le, E.~Chi, and Denny Zhou.
\newblock Self-consistency improves chain of thought reasoning in language models.
\newblock \emph{International Conference on Learning Representations}, 2022.
\newblock \doi{10.48550/arXiv.2203.11171}.

\bibitem[Wang et~al.(2023)Wang, Chen, Tan, and Guo]{tabi}
Yiding Wang, Kai Chen, Haisheng Tan, and Kun Guo.
\newblock Tabi: An efficient multi-level inference system for large language models.
\newblock In \emph{Proceedings of the Eighteenth European Conference on Computer Systems}, EuroSys '23, pp.\  233–248, New York, NY, USA, 2023. Association for Computing Machinery.
\newblock ISBN 9781450394871.
\newblock \doi{10.1145/3552326.3587438}.
\newblock URL \url{https://doi.org/10.1145/3552326.3587438}.

\bibitem[Wei et~al.(2022)Wei, Wang, Schuurmans, Bosma, Chi, Xia, Le, and Zhou]{cot}
Jason Wei, Xuezhi Wang, Dale Schuurmans, Maarten Bosma, E.~Chi, F.~Xia, Quoc Le, and Denny Zhou.
\newblock Chain-of-thought prompting elicits reasoning in large language models.
\newblock \emph{Neural Information Processing Systems}, 2022.

\bibitem[Xu et~al.(2024)Xu, Sun, Zheng, Geng, Zhao, Feng, Tao, Lin, and Jiang]{xu2024wizardlm}
Can Xu, Qingfeng Sun, Kai Zheng, Xiubo Geng, Pu~Zhao, Jiazhan Feng, Chongyang Tao, Qingwei Lin, and Daxin Jiang.
\newblock Wizard{LM}: Empowering large pre-trained language models to follow complex instructions.
\newblock In \emph{The Twelfth International Conference on Learning Representations}, 2024.
\newblock URL \url{https://openreview.net/forum?id=CfXh93NDgH}.

\bibitem[Yang et~al.(2018)Yang, Qi, Zhang, Bengio, Cohen, Salakhutdinov, and Manning]{yang2018hotpotqa}
Zhilin Yang, Peng Qi, Saizheng Zhang, Yoshua Bengio, William~W. Cohen, R.~Salakhutdinov, and Christopher~D. Manning.
\newblock Hotpotqa: A dataset for diverse, explainable multi-hop question answering.
\newblock \emph{Conference on Empirical Methods in Natural Language Processing}, 2018.
\newblock \doi{10.18653/v1/D18-1259}.

\bibitem[Yao et~al.(2023)Yao, Yu, Zhao, Shafran, Griffiths, Cao, and Narasimhan]{tot}
Shunyu Yao, Dian Yu, Jeffrey Zhao, Izhak Shafran, T.~Griffiths, Yuan Cao, and Karthik Narasimhan.
\newblock Tree of thoughts: Deliberate problem solving with large language models.
\newblock \emph{Neural Information Processing Systems}, 2023.
\newblock \doi{10.48550/arXiv.2305.10601}.

\bibitem[Yue et~al.(2023)Yue, Zhao, Zhang, Du, and Yao]{llmcascades}
Murong Yue, Jie Zhao, Min Zhang, Liang Du, and Ziyu Yao.
\newblock Large language model cascades with mixture of thoughts representations for cost-efficient reasoning.
\newblock 2023.

\bibitem[Zaharia et~al.(2024)Zaharia, Khattab, Chen, Davis, Miller, Potts, Zou, Carbin, Frankle, Rao, and Ghodsi]{compound-ai-blog}
Matei Zaharia, Omar Khattab, Lingjiao Chen, Jared~Quincy Davis, Heather Miller, Chris Potts, James Zou, Michael Carbin, Jonathan Frankle, Naveen Rao, and Ali Ghodsi.
\newblock The shift from models to compound ai systems.
\newblock \url{https://bair.berkeley.edu/blog/2024/02/18/compound-ai-systems/}, 2024.

\bibitem[Zellers et~al.(2019)Zellers, Holtzman, Bisk, Farhadi, and Choi]{hellaswag}
Rowan Zellers, Ari Holtzman, Yonatan Bisk, Ali Farhadi, and Yejin Choi.
\newblock Hellaswag: Can a machine really finish your sentence?
\newblock \emph{Annual Meeting of the Association for Computational Linguistics}, 2019.
\newblock \doi{10.18653/v1/P19-1472}.

\bibitem[Zheng et~al.(2024)Zheng, Zhou, Meng, Zhou, and Huang]{zheng2024large}
Chujie Zheng, Hao Zhou, Fandong Meng, Jie Zhou, and Minlie Huang.
\newblock Large language models are not robust multiple choice selectors.
\newblock In \emph{The Twelfth International Conference on Learning Representations}, 2024.
\newblock URL \url{https://openreview.net/forum?id=shr9PXz7T0}.

\bibitem[Zheng et~al.(2023{\natexlab{a}})Zheng, Chiang, Sheng, Zhuang, Wu, Zhuang, Lin, Li, Li, Xing, et~al.]{zheng2023judging}
Lianmin Zheng, Wei-Lin Chiang, Ying Sheng, Siyuan Zhuang, Zhanghao Wu, Yonghao Zhuang, Zi~Lin, Zhuohan Li, Dacheng Li, Eric Xing, et~al.
\newblock Judging llm-as-a-judge with mt-bench and chatbot arena.
\newblock \emph{arXiv preprint arXiv:2306.05685}, 2023{\natexlab{a}}.

\bibitem[Zheng et~al.(2023{\natexlab{b}})Zheng, Chiang, Sheng, Zhuang, Wu, Zhuang, Lin, Li, Li, Xing, Zhang, Gonzalez, and Stoica]{mtbench}
Lianmin Zheng, Wei-Lin Chiang, Ying Sheng, Siyuan Zhuang, Zhanghao Wu, Yonghao Zhuang, Zi~Lin, Zhuohan Li, Dacheng Li, Eric~P. Xing, Hao Zhang, Joseph~E. Gonzalez, and Ion Stoica.
\newblock Judging llm-as-a-judge with mt-bench and chatbot arena.
\newblock 2023{\natexlab{b}}.

\bibitem[Zhou et~al.(2023)Zhou, Schärli, Hou, Wei, Scales, Wang, Schuurmans, Cui, Bousquet, Le, and Chi]{leasttomost}
Denny Zhou, Nathanael Schärli, Le~Hou, Jason Wei, Nathan Scales, Xuezhi Wang, Dale Schuurmans, Claire Cui, Olivier Bousquet, Quoc Le, and Ed~Chi.
\newblock Least-to-most prompting enables complex reasoning in large language models.
\newblock 2023.

\end{thebibliography}
\bibliographystyle{iclr2024_conference}

\newpage
\onecolumn
\appendix
\section{Additional Dataset Details}
\subsection{Model Details \& Cost Estimation}
\label{appendix:model_info}
For all proprietary models, we calculate the cost of input and output results based on their API pricing. For open-source models, we utilize Together AI~\footnote{https://www.together.ai/pricing} to obtain results and reference costs. For the RAG experiment, we refer to the API pricing of You.com~\footnote{https://api.you.com/} and Perplexity~\footnote{https://docs.perplexity.ai/docs/pricing} for cost estimation.



\subsection{Dataset Details}
\label{appendix:dataset_details}
\textbf{MMLU}~\cite{mmlu}: A benchmark that measures the knowledge acquired by models during pretraining and evaluates models in zero-shot and few-shot settings across 57 tasks, testing both knowledge and reasoning on different fields of human knowledge.

\textbf{Hellaswag}~\cite{hellaswag}: This dataset challenges models to pick the best ending choice for a given sentence. It uses Adversarial Filtering(AF) to create a Goldilocks zone of complexity, wherein generations are largely nonsensical to humans but always make models struggle.

\textbf{GSM8K}~\cite{gsm8k}: A dataset of diverse grade school math word problems, testing a model’s ability to perform multi-step mathematical reasoning.

\textbf{ARC Challenge}\cite{arcc} A rigorous question answering dataset, ARC-Challenge includes complex, different grade-school level questions that require reasoning beyond simple retrieval, testing the true comprehension capabilities of models. Arc Challenge dataset contains those that both a retrieval and a co-occurrence method fail to answer correctly)

\textbf{Winogrande}~\cite{winogrande}: A large-scale and increased harness dataset inspired by the original Winograd Schema Challenge(WSC)~\cite{winograd} tests models on their ability to resolve pronoun ambiguity and their ability to understand the context with commonsense knowledge.

\textbf{MBPP}~\cite{mbpp}: The benchmark is designed to be solvable by entry-level programmers, covering programming fundamentals, standard library functionality, etc. Each problem comprises a task description, code solution, and 3 automated test cases.

\textbf{MT-Bench}~\cite{mtbench}: This dataset contains 3.3K expert-level pairwise human preferences for model responses generated by 6 models in response to 80 MT-bench questions, multi-run QA. The 6 models are GPT-4, GPT-3.5\footnote{https://openai.com/blog/chatgpt}, Claude-v1, Vicuna-13B~\cite{zheng2023judging}, Alpaca-13B~\cite{taori2023stanford}, and LLaMA-13B~\cite{touvron2023llama}. The annotators are mostly graduate students with expertise in the topic areas of each of the questions. In this work, we only used the 80 questions to generate model responses for~\dataset.

\subsection{More Details on Dataset Construction}
\label{appendix:data_entry}
Each sample in the benchmark dataset will have the following attributes:
\begin{itemize}
    \setlength{\itemsep}{0em}
    \item $sample\_id$: contain the information about the name of the sub-task, the split of dataset, and the index of the data in that dataset. Example: \textbf{mmlu-astronomy.val.5}
    \item $model\_name$: the model used to perform inference for this sample. Example: \textbf{GPT-4}
    \item $eval\_name$: the source data from which this specific sample comes. Example: \textbf{hellaswag.dev.v0}
    \item $prompt$: prompt sentence. Example: \textbf{The following are multiple-choice questions...}
    \item $model\_response$: Model's output. Example: \textbf{The answer is A)}
    \item $performance$: the result compared to the true label. Example: \textbf{True/False}
    \item $cost$: for proprietary model, we use API cost to calculate; for open source model, we use Together AI\footnote{https://www.together.ai/} to call the model and use their cost as reference. Example: \textbf{0.00019}
    \item $true\_label$: the true label or gold response for this prompt. Example: \textbf{True/False}
\end{itemize}

\subsection{Evaluation Metrics}
\label{appendix:evaluation_metric}
We will perform 5-shot inference on MMLU, HellaSwag, GSM8K, ARC Challenge, Winogrande, and 0-shot inference on MBPP, MT-Bench, and RAG.

For the datasets \textbf{MMLU}, \textbf{HellaSwag}, \textbf{GSM8K}, \textbf{ARC Challenge}, and \textbf{Winogrande}, we use the exact match method to compute the final results. In contrast, for \textbf{MBPP}, \textbf{MT-Bench}, and \textbf{RAG}, we use GPT-4 for answer evaluation. Results categorized as False/True are converted to a binary 0/1 format. In cases where the results are based on ratings, we normalize all outcomes to a [0, 1] scale.

\subsection{Individual Dataset Result}
The \dataset~pilot study result has been shown in \fig{fig:pilot}. We provide the breakdown of each dataset in \fig{appendix:more_dataset}. Additionally, we list the accuracies and costs for each individual model and the \textit{Oracle} router in \tbl{tab:all_res}.

\begin{figure*}[h] 
    \centering
    \includegraphics[width=1\textwidth]{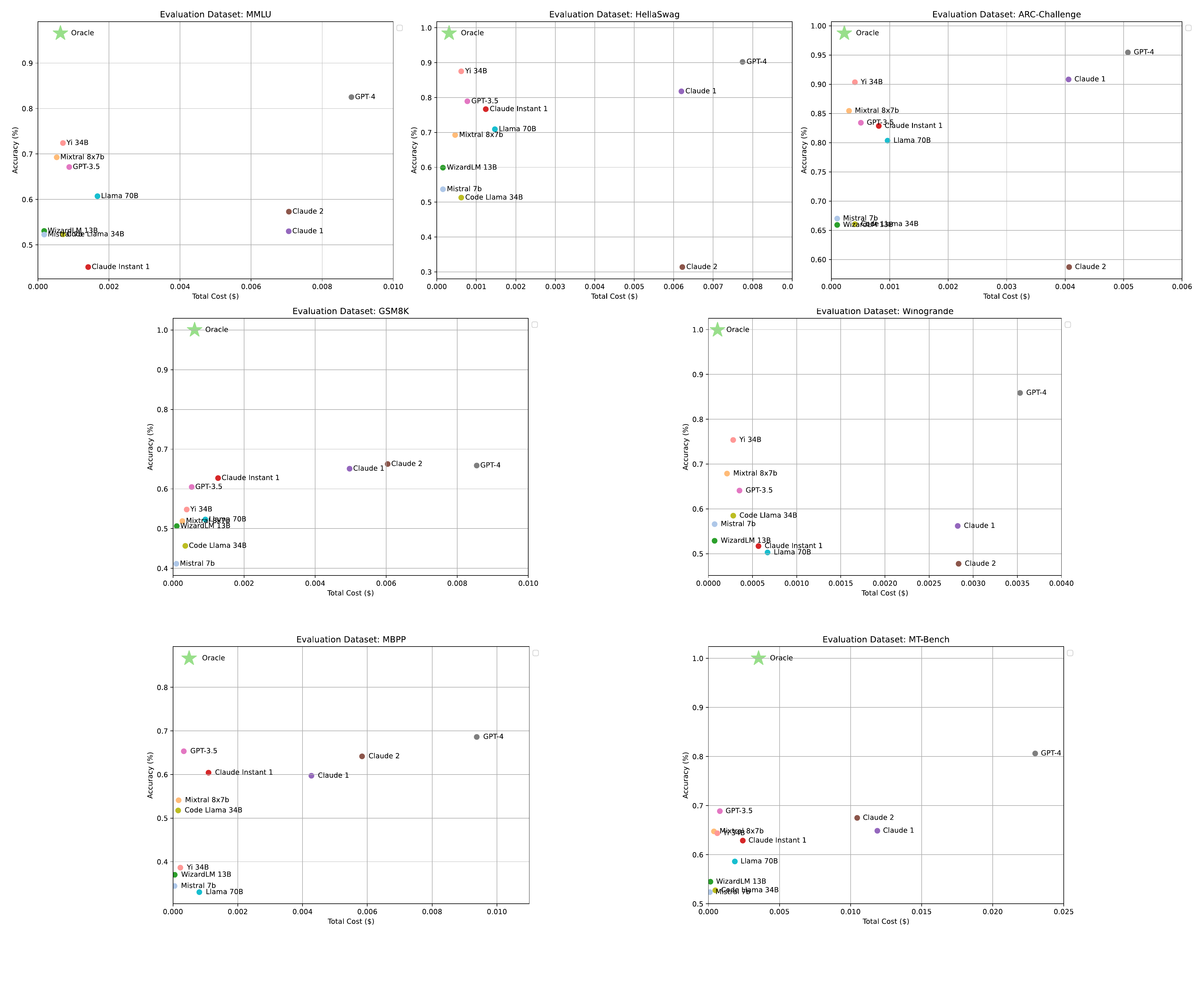}
    \caption{Accuracy vs Total cost of each LLM on each sub dataset in \dataset.}
    \label{appendix:more_dataset}
\end{figure*}

\begin{table*}
\caption{Individual models and the \textit{Oracle} results on the seven datasets. 
}\label{tab:all_res}
\vspace{1mm}
\centering
\scriptsize{
\setlength{\tabcolsep}{5pt}{
\begin{tabular}{c|cc|cc|cc|cc|cc|cc|cc}
        \toprule
        \multirow{2}{*}{\textbf{Method}} & \multicolumn{2}{c|}{\textbf{MMLU}}  & \multicolumn{2}{c|}{\textbf{MT-Bench}}  & \multicolumn{2}{c|}{\textbf{MBPP}}  & \multicolumn{2}{c|}{\textbf{HellaSwag}}  & \multicolumn{2}{c|}{\textbf{Winogrande}} & \multicolumn{2}{c|}{\textbf{GSM8k}} & \multicolumn{2}{c}{\textbf{ARC}}\\
          & Perf$\uparrow$ & Cost$\downarrow$ & Perf$\uparrow$ & Cost$\downarrow$  & Perf$\uparrow$ & Cost$\downarrow$  & Perf$\uparrow$ & Cost$\downarrow$ & Perf$\uparrow$ & Cost$\downarrow$ & Perf$\uparrow$ & Cost$\downarrow$ & Perf$\uparrow$ & Cost$\downarrow$ \\
        \midrule
        \midrule        
        WizardLM 13B & 0.568 & 0.122  & 0.796 & 0.006   & 0.364  & 0.011  & 0.636 & 0.727  & 0.512  &   0.040 &  0.510 & 0.354 & 0.660 & 0.068 \\
        Mistral 7B & 0.562 & 0.081  & 0.779   & 0.003  & 0.349  & 0.006  & 0.541 & 0.485  & 0.562  & 0.027  & 0.409  &  0.210 & 0.642 & 0.046 \\
        Mixtral 8x7B & 0.733 & 0.245  & 0.921   & 0.012  & 0.573  & 0.023  & 0.707 & 1.455  & 0.677  & 0.081  & 0.515  & 0.594 & 0.844 & 0.137 \\
        Code Llama 34B & 0.569 & 0.317  & 0.796   & 0.015  & 0.465  & 0.021  & 0.525 & 1.882  & 0.617  & 0.104  & 0.462  & 0.752 & 0.644 & 0.177 \\
        Yi 34B & 0.743 & 0.326  &  0.938  & 0.018  & 0.333  & 0.031  & 0.931  & 1.938  & 0.748  & 0.107  & 0.552  & 0.867 & 0.882 &  0.182 \\
        GPT-3.5 & 0.720 &  0.408 &  0.908  & 0.026  &  0.651 & 0.044  & 0.816 & 2.426  & 0.630  & 0.134  & 0.601  & 1.170 & 0.855 & 0.228 \\
        Claude Instant V1 & 0.384 & 0.327  & 0.863   & 0.030 & 0.550  & 0.064  & 0.801 & 1.943  & 0.512  &  0.108 &  0.626 & 1.300 & 0.821 & 0.183 \\
        Llama 70B & 0.647 &  0.367 & 0.854   & 0.022  & 0.302  & 0.039  & 0.736 & 2.183  & 0.504  &  0.121 &  0.529 & 0.870 & 0.794 & 0.205 \\
        Claude V1 & 0.475 & 3.269  & 0.938   & 0.361  & 0.527  & 0.607  & 0.841 & 19.43  & 0.570  & 1.077  & 0.653  & 11.09 & 0.889 & 1.829 \\
        Claude V2 & 0.619 & 3.270  & 0.854   & 0.277  & 0.605  & 0.770  & 0.421 & 19.50  & 0.446  &  1.081 & 0.664  & 13.49 & 0.546 & 1.833 \\
        GPT-4 & 0.828 &  4.086 & 0.971   & 0.721  & 0.682  & 1.235  & 0.923 & 24.29  & 0.858  & 1.346  & 0.654  & 19.08 & 0.921 & 2.286 \\
        \midrule
        \hc \textit{Oracle}  & \textbf{0.957}  & 0.297  & \textbf{0.996}  & 0.052  & \textbf{0.899}  & 0.041 & \textbf{0.994} & 0.860  & \textbf{1.0}  &  0.042 &  \textbf{0.748} & 1.282 & \textbf{0.977} & 0.091 \\
        \bottomrule
\end{tabular}
}
\vspace{-5mm}
}
\end{table*}

\section{Extended Experimental Settings}
We provide the hyperparameters of MLP and KNN routers in this section. 

The KNN routers have two main hyperparameters that were tested in this paper. The number of neighbors, and the embedding model for the prompts. All KNN routers used cosine similarity as the distance metric, and used either 5, 10, or 40 neighbors. The embedding models were taken from the default SentenceTransformers library \cite{reimers-2019-sentence-bert}, and are one of all-MiniLM-L12-v2, all-mpnet-base-v2, or all-distilroberta-v1. The best-performing hyperparameters for the KNN router were with 40 neighbors, and the all-MiniLM-L12-v2 embedding model.

In MLP routers, the models have either one or two hidden layers, with each layer having 100 neurons, and the ReLU activation function was applied. The learning rate was kept constant at 0.001, and the models took in embeddings from one of all-MiniLM-L12-v2, all-mpnet-base-v2, or all-distilroberta-v1. The best MLP router had two hidden layers of 100 neurons each, and used the all-MiniLM-L12-v2 embedding model.

\section{Issues with Overly-aligned Models}
\label{appendix:refuse}
Some models exhibit reluctance in responding to certain inputs, often replying with statements like "I do not understand..." or "I am not sure about...". We have identified two primary reasons for models' refusal to respond:

\myparagraph{Insufficient Context Perception} Despite being provided with enough context, these models perceive the information as inadequate. Our hypothesis is that the models' capabilities might not be robust enough to generate answers or perform tasks effectively under these conditions. A potential remedy is to modify the prompting strategy to encourage output generation.

\myparagraph{Uncertainty Avoidance} Some models appear to be fine-tuned to function as 'safe' assistants, refraining from providing responses when they lack certainty. This cautious approach likely aims to prevent potential errors stemming from uncertain answers. Claude 2 exhibits this behavior most frequently.

LLMs have been known to have such kind of issues as documented in various previous studies~\cite{zheng2024large, alzahrani2024benchmarks}. It is essential to apply methods that can make LLM outputs in a more controllable and structural way and automatically optimize their quality~\cite{khattab2024dspy, singhvi2023dspy} when routing, which warrants further exploration in future research.

\section{Full Cascading Routers Results}
Here are we provide the rest cascading routers results on ARC-Challenge, MT-Bench, and HellaSwag.

\begin{figure*}[!ht]
    \centering
    \vspace{-5pt}
    \includegraphics[width=1.0\linewidth]{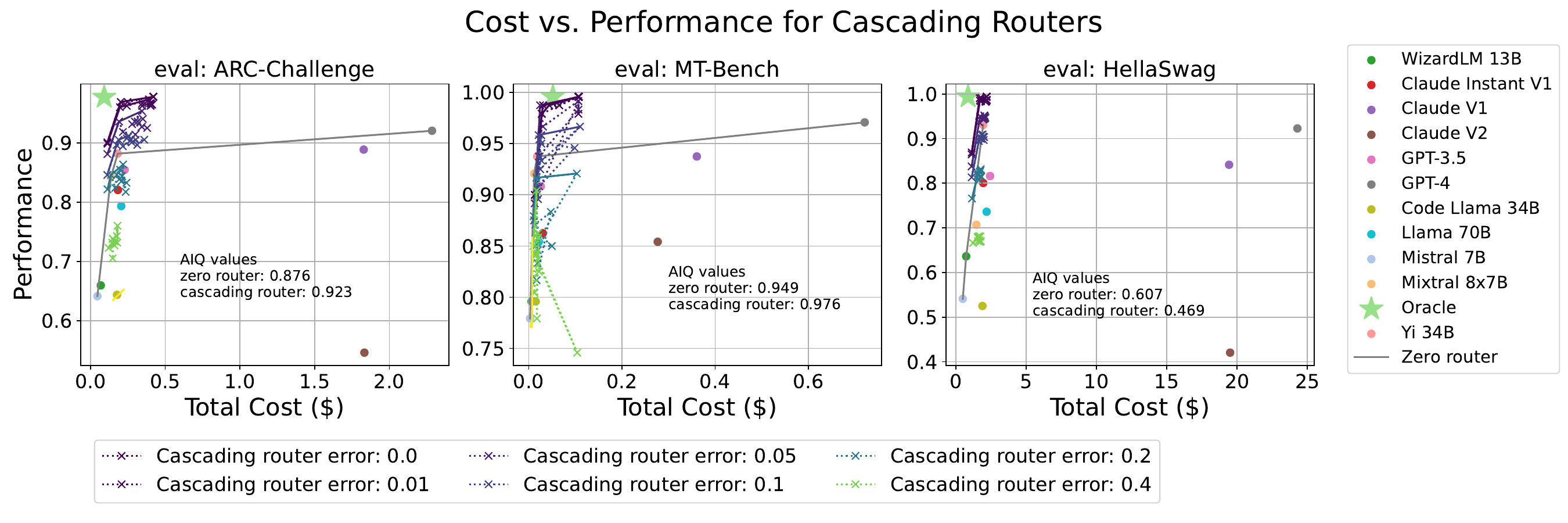}
    \vspace{-25pt}
    \caption{Total Cost vs Performance for eleven models and cascading routers on ARC-Challenge, MT-Bench, and HellaSwag. Different error rates are tested, and the AIQ value is computed for Zero Router and zero error rate cascading router. The solid lines represent non-decreasing convex hull and the dotted line represents points with increasing maximum cost parameter.}
    \label{cas_res_append}
    \vspace{-5pt}
\end{figure*}

\section{Training Data Distribution}
\begin{figure*}[!ht]
    \centering
    \vspace{-5pt}
    \includegraphics[width=1.0\linewidth]{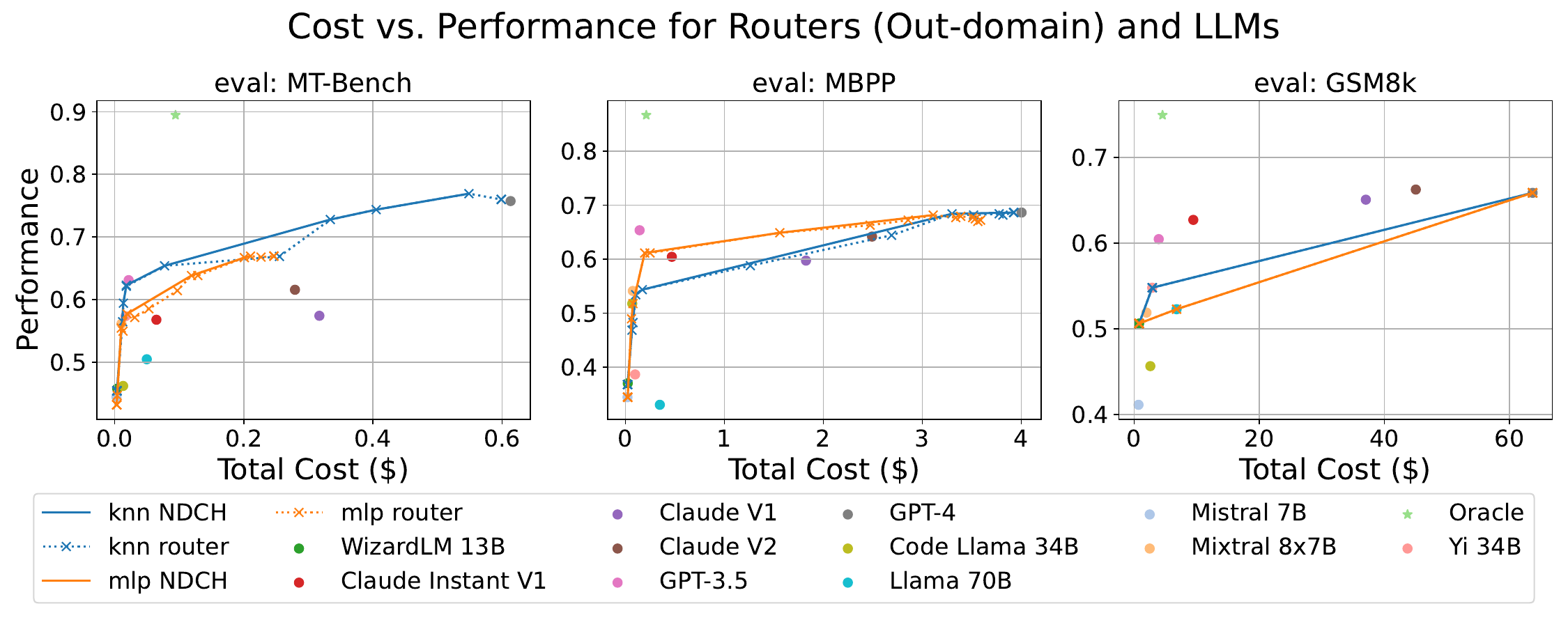}
    \vspace{-25pt}
    \caption{Total Cost vs Performance for eleven models and KNN, MLP routers on MT-Bench, MBPP, GSM8K. NDCH stands for non-decreasing convex hull}
    \label{results_od}
    \vspace{-5pt}
\end{figure*}

We also conduct Out-domain experiments where we train on held-out tasks in \dataset~for each dataset and evaluate on MT-Bench, MBPP and GSM8K in \fig{results_od}. 

\end{document}